\documentclass[article,5p, times, twocolumn]{elsarticle}

\usepackage{lineno,hyperref}
\modulolinenumbers[5]

\journal{Image and Vision Computing}

\usepackage{amsthm}
\usepackage{amsmath}
\usepackage{enumerate}
\usepackage{graphics}
\usepackage{multirow}
\usepackage{booktabs}
\usepackage{subfigure} 
\usepackage{hhline}

\usepackage{algorithm}
\usepackage[noend]{algpseudocode}
\usepackage{centernot} 

\makeatletter
\def\BState{\State\hskip-\ALG@thistlm}
\makeatother

\usepackage[dvipsnames]{xcolor}
\definecolor{mypink3}{cmyk}{0, 0.7808, 0.4429, 0.1412}

\newcommand{\etal}{\textit{et al}.}
\newcommand{\ie}{\textit{i}.\textit{e}.}

\newcommand{\eg}{\textit{e}.\textit{g}.}
\newcommand{\Wlog}{\textit{w}.\textit{l}.\textit{o}.\textit{g}.}

\biboptions{sort&compress}

\begin{document}

\begin{frontmatter}


\title{Introducing the structural bases of typicality effects in deep learning}


\author{Omar Vidal Pino\corref{mycorrespondingauthor}} 
\ead{ovidalp@dcc.ufmg.br}
\cortext[mycorrespondingauthor]{Corresponding author.}

\author{Erickson~R.~Nascimento} 
\ead{erickson@dcc.ufmg.br}

\author{Mario~F.~M.~Campos}
\ead{mario@dcc.ufmg.br}

\address{Computer Vision and Robotics Laboratory, Computer Science Department, Universidade Federal de Minas Gerais,\newline Belo Horizonte, 31270-010, Brazil}







\begin{abstract}
In this paper, we hypothesize that the effects of the degree of typicality in natural semantic categories can be generated based on the structure of artificial categories learned with deep learning models. Motivated by the human approach to representing natural semantic categories and based on the foundations of Prototype Theory, we propose a novel Computational Prototype Model (CPM) to represent the internal structure of semantic categories. Unlike other prototype learning approaches, our mathematical framework proposes a first approach to provide deep neural networks with the ability to model abstract semantic concepts such as category central semantic meaning (semantic prototype), typicality degree of object's image,  and family resemblance relationship. We proposed several methodologies based on the typicality's concept to evaluate our CPM model in image semantic processing tasks such as image classification, a global semantic description of images,  and transfer learning. Our experiments on different image datasets, such as ImageNet and Coco, showed that our approach might be an admissible proposition in the effort to endow machines with greater power of abstraction for the semantic representation of objects' categories.


\end{abstract}

\end{frontmatter}


\section{Introduction}
\label{sec:introduction}

Memory is one of the most amazing faculties of the human being and construed as the brain's ability to code, store, and retrieve information~\cite{atkinson1968,tulving2007coding, yee2018semantic,netto2015association}. For decades, understanding and simulating the basis of human learning,  cognition processing, and its perception and vision system has been the motivation of the machine intelligence field. In recent years, pattern recognition methods with an impressive performance for some specific tasks related to image interpretation have been developed in the Computer Vision and Image Processing fields. However, these methods still lack in others capabilities compared to human proficiency. Image semantic understanding is influenced by how the features of image basic components~(\eg, objects) are semantically represented and how the semantic relationships between these basic components are constructed~\cite{guo2016deep}.  Knowledge extraction models (high-level vision processes) from images are highly influenced by the methods used to detect, extract, and represent the image's relevant semantic information.


The advent of Convolutional Neural Networks (CNN) outperformed the traditional methods~\cite{lowe2004SIFT, bay2008SURF} used for image feature representation, and CNN-methods are the leading approaches in semantic image processing tasks such as object recognition~\cite{simonyan2014very}, semantic segmentation~\cite{shelhamere2017darrellt}, object description~\cite{Li_2020_CVPR}, semantic correspondence~\cite{Rocco18}, etc. Although state-of-the-art CNN methods have achieved remarkable results,  there are still many challenges to attain the discriminative power and the abstraction of human memory~(\eg, semantic memory~\cite{tulving2007coding, yee2018semantic}) to represent the semantic of visually acquired information. How to emulate the behavior of human memory in the representation of learned knowledge of objects’ features? How to extract and encode such features to encapsulate the representation of the meaning (or semantic representation) of a specific object? How to infer or ascribe semantics to objects? How to represent the image's meanings and its phenomena? The quest to answer some of these questions still occupies the investigation agenda of many researchers.

\begin{figure*}[t!]
	\begin{center}
		
		\includegraphics[width=0.88\linewidth]{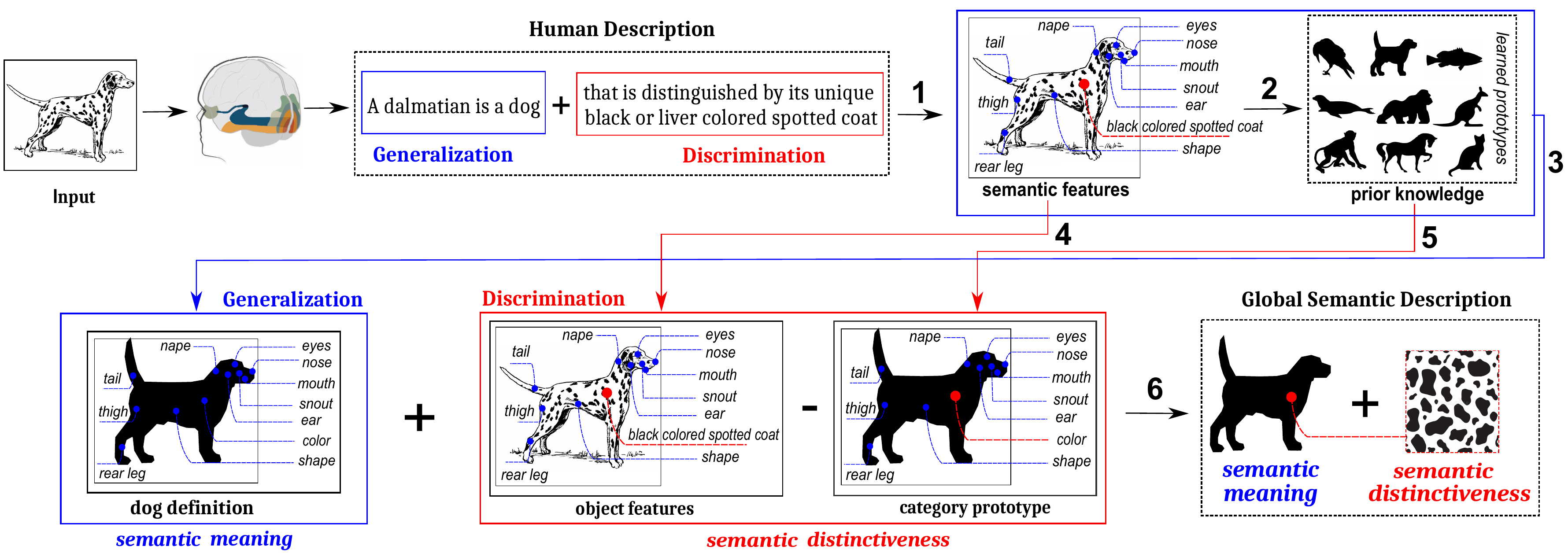}
	\end{center}
	 \vspace*{-\baselineskip}
	\caption{\textbf{Schematic of our prototype-based classification and description models}. 
	The human visual system can observe, categorize and build the semantic description of an object based on its most distinctive features within that object's category. We propose a prototype-based model to simulate this behavior through the pipeline composed of modules \textit{1)} to \textit{6)}.
		\textit{1)} features extraction; \textit{2)} recognition of object features; \textit{3)} prototypes-based classification; \textit{4)} object features; \textit{5)} central semantic meaning of a  category~(the category prototype); \textit{6)} our Global Semantic Description based on Prototypes.}
	\label{fig:motivation}
	
\end{figure*}


Object typicality effects are among these semantic phenomena that are difficult to capture, and they are still challenging for the image computing process. The typicality concept refers to the degree to which the objects under study are considered good examples of the category~\cite{rosch1973internal, rosch1975family}. For example, the pigeon is a typical case in the bird category since it has several representative features: it can fly, has feathers, beaks, lays eggs, and builds a nest. On the other hand, the penguin is an atypical member since it satisfies only some features but not all. A glance is enough for human beings to perform this type of semantic ranking within the category. In contrast, once objects belong to the same category, machines still lack the ability to capture this semantic phenomenon.

The argument that category membership is a matter of degree came from cognitive psychology with the seminal studies of Rosch and colleagues~\cite{rosch1973internal,rosch1975family,rosch1975cognitive,rosch1976structural,rosch1978principles}, who referred to membership degree as \textit{typicality}. In her seminal work~\cite{rosch1975family}, Rosch introduced the concept of semantic prototype and presented an in-depth analysis of the internal semantic structure of the category. Rosch~\cite{rosch1975family} holds that the representation of category semantic meaning is related to the category prototype, particularly to those categories denoting natural objects. The Prototype Theory~\cite{rosch1973internal,rosch1975family,rosch1975cognitive,rosch1976structural,rosch1978principles,geeraerts2010theories} proposes that human beings think categories in terms of abstraction (prototypes), represented by typical category members. This theory also indicates that the successful execution of object classification and description tasks in the human brain is inherently related to the category prototype learned.

This paper relies on cognitive semantic studies related to the Prototype Theory to propose a new perspective to model the central semantic meaning of object categories: the prototype. Unlike other prototype learning approaches \cite{Ma2013,Ojeda2013,wohlhart2013optimizing, saleh2013object,Zhao2015,  jetley2015prototypical,saleh2016incorporating, Oyedotun2017, snell2017prototypical, drumond2017using, dong2018few, fort2018gaussian, Yang2018, allen2019infinite,angelov2020towards,  xiao2020tdapnet, garnot2021leveraging}, we use our prototype representation to capture the concepts of typicality and category membership degree of object's images. Our proposal considers the typicality concept from cognitive psychology, assuming that it is possible to obtain a more natural and interpretable representation of the semantics of the object's image. Specifically, we propose a mathematical framework that endeavors to represent the semantic definition of an object's categories and, consequently, capture the phenomena of the object's typicality. To evaluate our proposal in real-world tasks, we also propose a procedure to introduce our prototype's semantic representation and our typicality measure in the global semantic description of the object's images. Furthermore, we also propose a CNN-layer architecture to evaluate our proposal in classification and transfer learning tasks. Figure~\ref{fig:motivation} shows the intuition and our basic conceptual steps to apply our framework to classification and description models.

Prototype learning is a representative approach of pattern recognition methods. It has been used in such image processing tasks as Face Recognition~\cite{Ma2013,Oyedotun2017}, Image Segmentation~\cite{Ojeda2013,dong2018few}, Static Hand Gesture~\cite{Oyedotun2017}, Few-Shot Learning~\cite{jetley2015prototypical, snell2017prototypical, fort2018gaussian, allen2019infinite}, Clustering~\cite{Zhao2015}, Robust Image Classification~\cite{wohlhart2013optimizing, saleh2013object,saleh2016incorporating, Yang2018, xiao2020tdapnet, garnot2021leveraging}, CNN Interpretation~\cite{drumond2017using, angelov2020towards}, etc. Even though these works proposed a wide variety of methods for prototype learning, most of the approaches focus on using prototype learning to improve the performance/robustness of a specific task. As far as we know, little attention is paid to the use of the prototype to capture other semantic properties of the object image, such as its typicality; further, if we consider that the prototype is based on the notion of typicality~\cite{rosch1975family, rosch1975cognitive,rosch1978principles, geeraerts2010theories}. 

Introducing typicality into image processing can increase the generalization power of pattern recognition models. This assumption can be supported -- on the one hand -- due to its theoretical foundations: Rosch's experiments~\cite{rosch1978principles} showed that when humans learn a category by looking at its most typical samples, they can better recognize new members. On the other hand, some authors~\cite{saleh2013object,saleh2016incorporating} showed that deep learning models could not generalize atypical images that are substantially different from training images. When a typicality measure is involved in the learning process, we can improve the image classification task~\cite{saleh2013object, saleh2016incorporating}. Moreover, involving the typicality learning in the category's learning process would allow the machines to categorize the images and know their degree of belonging (is it a typical, atypical, or border-image?), a type of semantic representation of the object's image categories that are only achievable by human beings.

Our experimental results show that our mathematical framework allows us to interpret possible semantic associations between members within the category’s internal structure. Results also indicate that our method could establish a relationship between the proposed typicality measure and the representativeness of the object’s image. Experiments on ImageNet~\cite{ILSVRC15} and Coco~\cite{lin2014microsoft} datasets present that good performance can be achieved in real-world tasks while capturing other semantic properties of the object's image, like typicality.

\section{\textbf{Related Work}}
\label{sec:related_work}
 
\paragraph{Prototype Theory} 
The Prototype Theory~\cite{rosch1973internal,rosch1975family,rosch1975cognitive,rosch1976structural,rosch1978principles, geeraerts2010theories} analyzes the internal structure of semantic categories and proposes categorization based on the prototype. This Theory postulates that semantic categories are not homogeneous structures. According to experimental evidences~\cite{rosch1973internal,rosch1975family, rosch1975cognitive,rosch1976structural}, semantic categories should be considered heterogeneous structures, where their members and their respective characteristics do not have the same relevance within the category.

\begin{figure}[t!]
	\begin{center}
		\includegraphics[width=0.9\linewidth]{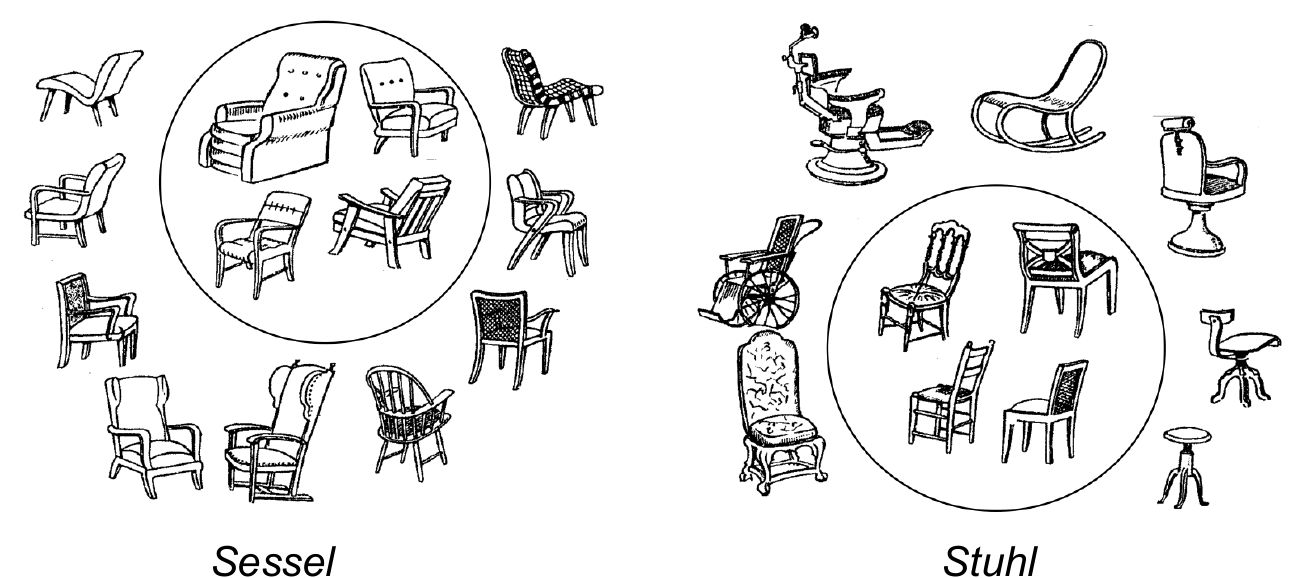}
	\end{center}
	\vspace*{-\baselineskip}
	\caption{\textbf{Category's prototypical organization}. The figure shows the \textit{Sessel} and~\textit{Stuhl} experiment conducted by Gipper~(Figure adapted from~\cite{geeraerts2010theories}). The experiment studies the meaning of German words \textit{Stuhl}~(chair) and \textit{Sessel}~(comfortable chair) and shows that within the \textit{chair} category, the category's internal organization (and central semantic meaning) can change according to the given criterion and object typicality.} 
	\label{fig:chairs_experiment}
\end{figure}

Rosch and colleagues~\cite{rosch1973internal,rosch1975family,rosch1975cognitive,rosch1976structural,rosch1978principles, geeraerts2010theories} argued that semantic categories are made up of good and bad examples, according to a given criterion. The most representative members (typical members), those that are evoked when thinking/or viewing a category, are the central members (focal cases) or prototypical members (best examples) around which the rest of the category members are organized; thus exhibiting a prototypical organization of the category. Figure~\ref{fig:chairs_experiment} shows an example of the \textit{prototypical organization} phenomenon. According to the authors~\cite{rosch1973internal,rosch1975family,rosch1975cognitive,rosch1976structural,rosch1978principles, geeraerts2010theories} the prototype is based on typicality, and all members belonging to the same category do not represent it in the same way, \ie, some members are more typical than others. Accordingly, there must be an internal \textit{family resemblance} among category members in each category and an external dissimilarity (low similarity)  with the members of the other categories.

The prototype was initially defined as the most representative and distinctive member of a category~\cite{rosch1973internal}, as it is the element that shares more features with the other category members and less with members of other categories. But one of the problems of defining the prototype as an element or as a \textit{prototype-object} is related to the decision of who is the category's prototype when two members are equally representative? Consequently, the prototype began to be defined as a \textit{prototype-cognitive entity}, specifically, as prototypicality effects~\cite{rosch1975family,rosch1975cognitive,geeraerts2010theories}.

The category prototype was formally defined as a cognitive entity as the clearly central of a given category~\cite{rosch1975family,rosch1975cognitive}~(for example, in Figure~\ref{fig:chairs_experiment} the prototype is defined by the typical members within the circle). The attributes of those central members are those that are structurally the most salient category properties.
Rosch's experiments~\cite{rosch1973internal,rosch1975family,rosch1975cognitive,rosch1976structural,rosch1978principles} showed that human beings store category knowledge as a semantic organization around the category's prototype. The categorization of an object is obtained based on the similarity of a new exemplar with one of the prototypes (cognitive abstraction) learned.

According to Geeraerts~\cite{geeraerts2010theories}, the concept
of prototypicality is in itself prototypically clustered based on one to four characteristics. The concepts of \textit{non-discreteness} and \textit{non-equality} (either on the \textit{intensional} or on the \textit{extensional} level) play a major distinctive role. Four characteristics are frequently mentioned as typical of prototypicality in semantic categories~\cite{rosch1975cognitive,geeraerts2010theories}: \textit{i)} categories exhibit degrees of typicality; not every member is equally represented in the category~(extensional non-equality); \textit{ii)} categories are blurred at the edges~(extensional non-discreteness); \textit{iii)} categories are clustering into family resemblance structure, \ie, the category semantic structure takes the form of a radial set of clustered and overlapping members~(intensional non-equality); and \textit{iv)} categories cannot be defined by a single set of criteria~(necessary and sufficient) attributes~(intensional non-discreteness). The prototypicality effects~(Table~\ref{table:prototypicality-effects}) surmise the importance of the distinction between the central and peripheral meaning of the object categories~\cite{geeraerts2010theories}.

\begin{table}[t!]
	\caption{Two-dimensional conceptual map of prototypicality effects according to Geeraerts~\cite{geeraerts2010theories}.}
	\label{table:prototypicality-effects}
	\resizebox{\columnwidth}{!}{%
		\begin{tabular}{lcc}
			\toprule
			
			& \textit{Extensional} & \textit{Intensional}\\
			\midrule
			\textit{non-equality} & Difference of typicality & Clustering into family\\{\scriptsize(salience effect,} & and
			membership salience & resemblances\\ {\scriptsize core/periphery)}&&\\
			\midrule
			\textit{non-discreteness} & Fuzziness at the edges,&Absence of necessary and\\{\scriptsize (demarcation }& membership uncertainty& sufficient definitions\\ {\scriptsize problems, flexibly)}&&\\
			\bottomrule
		\end{tabular}
	}
    \vspace*{-\baselineskip}
\end{table}

\paragraph{Prototype Learning}

Learning Vector Quantization~(LVQ) is a field sprung by the seminal work of Kohonen~\cite{kohonen1997learning}, in which the methods attempt to find optimal prototypes from labeled data. 
LVQ models partition the input space and assign each partition a set of prototypes~\cite{kohonen1997learning}. The classification of a new element is based on the proximity (similarity) with the learned prototypes. 
LVQ approach has been widely studied by many works~\cite{kohonen1997learning, seo2003soft}. It has many variations that normally differ in the proposal used for feature extraction (handcrafted features) and the approach used to prototypes construction/update. Works like~\cite{Yang2018,liu2001evaluation} presented a more detailed review of this family of prototype-based learning methods.

With the rise of deep neural networks, handcrafted features were replaced with CNN features in prototype learning, thus achieving end-to-end integration in deep networks and high precision and robustness in various image processing tasks. The differences between the great variety of existing approaches can be roughly grouped by: \textit{i)} the number of prototypes used to represent a category (1-per-class~\cite{Ojeda2013,dong2018few,jetley2015prototypical, snell2017prototypical, fort2018gaussian, wohlhart2013optimizing, garnot2021leveraging}, n-per-class~\cite{Oyedotun2017, allen2019infinite, Yang2018, xiao2020tdapnet, drumond2017using}, sparse~\cite{Ma2013}); \textit{ii)} the distance measure  (or measures combination) used to stand for the similarity between each instance-prototype pair (Euclidean distance~\cite{Ma2013,Oyedotun2017,Ojeda2013,snell2017prototypical, allen2019infinite, wohlhart2013optimizing,Yang2018, xiao2020tdapnet}, Mahalanobis distance~\cite{Ojeda2013,  wohlhart2013optimizing}, Co-variance distance~\cite{fort2018gaussian}, Cosine distance~\cite{snell2017prototypical},  Learned distance~\cite{dong2018few, drumond2017using}, Hand-designed distance~\cite{xiao2020tdapnet, garnot2021leveraging}); and  \textit{iii)} the approach used for prototype representation  (prototype-template image~\cite{Ma2013, jetley2015prototypical}, mean vector of embedded features~\cite{Oyedotun2017, dong2018few,snell2017prototypical, fort2018gaussian, allen2019infinite}, learned centroid vector~\cite{Ojeda2013,  wohlhart2013optimizing, Yang2018,garnot2021leveraging, drumond2017using},  learned CNN-tensor~\cite{xiao2020tdapnet}).

All these previous approaches improve the performance of a specific task and: 1) the prototype learning use all images of the training dataset (regardless of whether they are good or bad examples of the category); 2) although these methods assume that the categories are prototypical, they also assume that the categories are homogeneous (and consequently use similarity measures that do not take into account the relevance of the attributes for each category). Note that these two characteristics show that current prototype learning methods~\cite{Ma2013,angelov2020towards} do not consider the theoretical foundations established by Rosch~\cite{rosch1975family,rosch1975cognitive} to represent the semantics structure of prototypical categories; therefore, they are still unable to capture higher-level semantic phenomena, such as object's image typicality.
In the next section, we present a mathematical framework based on the foundations of Prototype Theory that presents a first approach to the use of prototypes to capture the object typicality.

\section{\textbf{Computational Prototype Model}}
\label{sec:CPM_model}



The semantic structure of a category~(\ie, core and peripheral meaning) is related to differences of typicality and membership salience among category members~(extensional non-equality in Table~\ref{table:prototypicality-effects})~\cite{rosch1978principles, geeraerts2010theories}. Rosch's experiments showed that the heterogeneous internal structure of natural semantic categories relates to the concepts of \textit{prototype}~(core meaning of the category), \textit{typicality}, and \textit{family resemblance}.
Family resemblance relationship~\cite{rosch1975family, rosch1976structural} consists of a set of items of the form AB, BC, CD, DE; \ie, each item has one or more attributes in common with one or more other items, but no attribute needs to be common to all items~\cite{rosch1975family}. The abstract nature of these semantic concepts has made it difficult to simulate them, even with the most powerful current techniques: deep learning.

Rosch's experiments also concluded that family resemblance is a function of the frequency (or learned attributes relevance) and the distribution of attributes\cite{rosch1976structural}. 
It is worth noting that the attributes relevance (category weights) and the distribution of the attributes (category features distribution) are, currently, characteristics that can be modeled with CNN models. In the following, we used Rosch's results as assumptions to model our Computational Prototype Model~(CPM).


\subsection{Semantic Prototype Representation}

In general, prototype learning methods~\cite{Ma2013,Ojeda2013,wohlhart2013optimizing, saleh2013object,Zhao2015,  jetley2015prototypical,saleh2016incorporating, Oyedotun2017, snell2017prototypical, drumond2017using, dong2018few, fort2018gaussian, Yang2018, allen2019infinite,angelov2020towards,  xiao2020tdapnet, garnot2021leveraging} represent the prototype as a centroid vector computed using all category members. In contrast to those proposals that assume the prototype as a centroid element (prototype-object), and based on Rosch’s prototype definition, we propose representing the prototype as a \textit{semantic entity} with a center and boundaries computed using \textit{only the typical members.}




\newdefinition{definition}{Definition}

Let $O$ be a set of objects  
and $C=\left\{{c_1, c_2,...,c_n}\right\}$~be the set of objects categories that partitions $ O $;  ${O_{c_i} = \left\{{o \in O: {\small category(o)} = c_i}\right\}}$ is the set of objects that share the same \textit{i}-th category $c_i \in C$, $\forall i = 1,...,n,$ and ${F =\left\{ {f_1, f_2,...,f_m}\right\}} $ is the set of features of an object.

 \begin{definition}{\textit{Semantic prototype}.}
We define a \textit{semantic prototype} as the central meaning of category  $c_i \in C$. Thus, the semantic prototype is given by the average and standard deviation of each features of \textit{all typical objects} within the  $c_i$-category along with a measure of the relevance of those features. Formally, the semantic prototype is represented by the tuple $ P_i = \left( {M_{i},\Sigma_{i},\Omega_{i}}\right), $ where $\:  \forall i = 1,...,n;\forall j = 1,...,m $ and  
	\begin{enumerate}[i)]
		\item $ M_i = \left[  {\mu_{i1}, \mu_{i2},...,\mu_{im}}\right] $ and $\mu_{ij}$ is the mean of j-$th$ feature considering only typical objects of $c_i$-category; 
		
		\item $ \Sigma_{i} = \left[ {\sigma_{i1}, \sigma_{i2},...,\sigma_{im}}\right] $ and $\sigma_{ij}$ is the standard deviation of j-$th$ feature considering only typical objects of $c_i$-category; 
		
		\item $ \Omega_{i} = \left[ {\omega_{i1}, \omega_{i2},...,\omega_{im}}\right] $ and $\omega_{ij}$ is the relevance value of j-$th$ feature for the category $c_i \in C$.
	\end{enumerate}

An \textit{abstract prototype} is defined as the ideal element or the most prototypical element of the $i$-th category; and it is given by the $ m $-dimensional vector $ M_i \in P_i$ composed of the expected value of the most salient features of $i$-th category since was computed using only typical members.
	\label{def:semantic_pttype}
 \end{definition}

Rosh and colleagues~\cite{rosch1973internal,rosch1975family,rosch1975cognitive,rosch1976structural,rosch1978principles} besides arguing that the prototype is the semantic nucleus of natural categories; also stated that from this nucleus, the categorical continuum could be characterized by two gradations:~\textit{i)} the relative importance evaluates each attribute it has for the category, and \textit{ii)} the relevance (or salience) of each category member coincides with the amount and type of features that the element presents.  In this way, it is possible to establish the prototypicality degree of a given element within the category~\cite{rosch1975family, rosch1975cognitive}. For example, within the color category, the weight and height attributes are not relevant (null relevance); conversely, within the category ``light objects'', the relevance of color and height attributes is null, and weight is very high. Consequently, the object relevance is evaluated based on its weight attribute. Note that this type of semantic distance relative to the semantic prototype cannot be modeled with similarity measures that assume that the categories' attributes are homogeneous (\eg, the Euclidean distance and other classical measures used in prototype learning).

\subsection{Semantic Distance}


Formal models of experimental psychology such as Prototype Model~\cite{homa1976category}, Multiplicative Prototype Model (MPM)~\cite{minda2002comparing}, and Generalized Context Model (GCM)~\cite{medin1978context, zaki2003prototype}  proposed measures of semantic distances between the stimulus that corresponds to Prototype Theory foundations. In Definition~\ref{def:obj_distance}, we present a semantic distance between objects as a measure of the family resemblance. Our proposal is a generalization of the psychological distance between two stimuli proposed in the GCM formal model.
Unlike the original formal Context Model~\cite{medin1978context}, we assume that object features (stimuli) are not binary values~($f_j \in \mathbb{R}$) and the relevance~($\omega_{ij}$) (or cost of attention) of each $j$-th unitary object feature is forced to be strictly positive, but has no upper limit, \ie,  $\sum_{j=1}^{m} \omega_{ij}\neq 1$.



\begin{definition}{\textit{Objects Dissimilarity.}}
	Let $ {o_1, o_2 \in O_{c_i}}$ be a representative objects of $ i $-th category $c_i \in C$ and $\emph{F}_{o_1}, \emph{F}_{o_2}$ the features of objects $o_1, o_2$ respectively. We defined the objects dissimilarity or the  objects distance between $o_1$ and $o_2$ as the semantic distance given by
	\begin{eqnarray} 
		\delta(o_1,o_2) = \sum_{j=1}^{m} \left| \omega_{ij}  \right| \times \left|f_j^1-f_j^2\right|,
		\label{eq:objects distance}
	\end{eqnarray} 
	where $\omega_{ij} \in \Omega_{i},$ $f_j^1 \in F_{o_1}$, $f_j^2 \in F_{o_2}$, and $\left|\cdot \right|$ is L1-norm, $\: \forall i= 1 \dots n;\,\forall j = 1 \dots m$.
	\label{def:obj_distance}
\end{definition}

 


\begin{definition}{\textit{Prototypical distance.}}
	Let $ {\emph{o} \in O_{c_i}}$ a representative object of $ i $-th category $c_i \in C$, $\emph{F}_o$ the features of object $\emph{o}$ and  $ P_i = \left( {M_{i},\Sigma_{i},\Omega_{i}}\right)$ the semantic prototype of $c_i$-category. We defined as \textit{prototypical distance}  between $\emph{o}$ and $P_i$ the semantic distance:
	\begin{eqnarray} 
		\delta(\emph{o},P_i) = \sum_{j=1}^{m} \left| \omega_{ij}\right| \times \left|f_j-\mu_{ij}\right|,
		\label{eq:pttype_distance}
	\end{eqnarray} 
	\noindent where $\omega_{ij} \in \Omega_{i},\,\mu_{ij} \in M_{i},$ and $f_j \in F_o\,;$ $ M_{i},\Omega_{i} \in P_i$ $\: \forall i= 1 \dots n;\,\forall j = 1 \dots m$.
	\label{def:pttype_distance}
\end{definition}


The proposed prototypical distance is a generalization of the semantic distance of the MPM formal model~\cite{minda2002comparing}.  Different from MPM model assumptions, we assumed that prototype features are the features of the ideal member (abstract prototype) of $i$-th category($ M_{i} \in P_i$). Note that our prototypical distance (Def.~\ref{def:pttype_distance}) is a specific case of our dissimilarity measure between objects (Def.~\ref{def:obj_distance}),  where one of the elements is the abstract prototype.

Since the distance function $\delta : {F}_{c_i} \times {F}_{c_i} \to \mathbb{R}^{+}$ (${F}_{c_i}$ is a non empty set of all objects features of category $c_i \in C$) satisfies the axioms of non-negativity, identity of indiscernible, symmetry and triangle inequality;~$\delta $ is a metric in the features domain ${F}_{c_i}$. Consequently, $({F}_{c_i},\delta)$ is a \textit{metric space} or features metric space.  Notice that $({F}_{c_i},\delta)$ is a measurable space (see proof in the supplementary material).



Since $({F}_{c_i},\delta)$ is a measurable space, we can use the generalization of Chebyshev's inequality to define the boundary of our semantic prototype representation. 
Stellato~\etal~\cite{stellato2017multivariate} approached the problem of formulating an empirical Chebyshev inequality given $N$ i.i.d. samples from an unknown distribution $\Pr$ and their empirical mean $\mu_{N}$ and standard deviation $\sigma_{N}$. The authors derive a Chebyshev inequality bound with respect to the $ (N + 1) $-th sample. The Multivariate Chebyshev inequality~\cite{stellato2017multivariate}  can define the boundary for an ellipsoidal set centered at the mean.
Consequently, we construct a confidence ellipsoidal set from the sample mean~($M_i$) and std~($\Sigma_{i}$) of only typical objects samples by computing a threshold vector $\vec{\lambda_{i}}$.  Let $E \subseteq {F}_{c_i}$ be a set of features extracted for only typical objects of $c_i$-category, and $F_{o} \subseteq E$ the features of a typical object $o \in O_{c_i}$. We weakly define as \textit{edges} of our semantic prototype the threshold vector $ \vec{\lambda_{i}} = \left[ {\lambda_{i1}, \lambda_{i2},\dots,\lambda_{im}}\right]$ that meets the expression:
\begin{eqnarray} 
	\Pr(|f_j-\mu_{ij}|\geq \lambda_{ij} \sigma_{ij}) \leq \min \left(1, \frac{1}{\lambda_{ij}^2}\right) ,
	\label{eq:semantic_edges}
\end{eqnarray} 
\noindent where $f_j \in F_{o}$, $\mu_{ij} \in M_{i}$, and $\sigma_{ij} \in \Sigma_{i}$. 

\begin{figure}[t!]
	\begin{center}
		\includegraphics[width=0.8\linewidth]{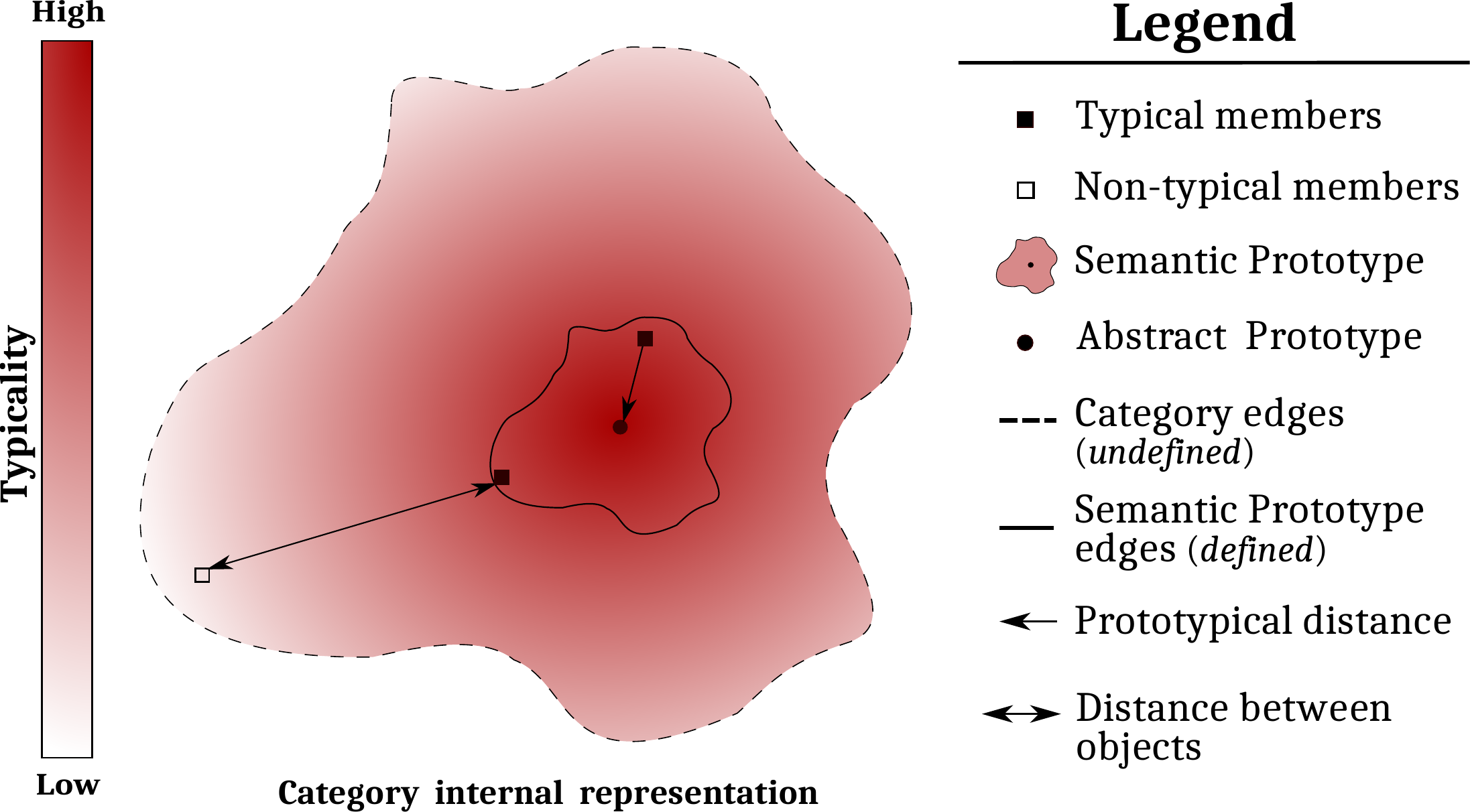}
	\end{center}
	\vspace*{-\baselineskip}
	\caption{\textbf{Category internal structure}. The expected semantic representation of a category's internal structure. The diagram also shows the key definitions and constraints of our Computational Prototype Model.}
	\label{fig:our_CPM_model}
\end{figure}

Figure~\ref{fig:our_CPM_model} shows the theoretical representation of category internal structure based on our CPM model. Our approach considers important concepts of the Prototype Theory, namely:  \textit{i)} our semantic prototype encoding is computed using only typical samples; \textit{ii)} category prototype edges are well defined; 
\textit{iii)} category edges provide a fuzzy definition because our semantic prototype is not computed with all category elements; \textit{iv)} objects representativeness degree~(typicality) within the category is simulated with our prototypical distance, and \textit{v)} family resemblance relationship is simulated with our objects dissimilarity measure.


\subsection{Prototype Construction}
\label{sec:prototype}
In this paper, we exploit the capability of CNNs in image semantic processing and classification tasks and use them as backbone of the components of our CPM framework.  We assume as category attribute distribution the features distribution~(${F}_{c_i}$) composited with object's-images features extracted from the last dense layer (before the softmax layer) of a CNN model;  object's images that belong to the same category. In addition, we assume as relevance ($ \Omega_{i}$) of the category's attributes those category's weights learned by the softmax layer.

\begin{figure*}[t!]
	\begin{center}
		\includegraphics[width=0.8\linewidth]{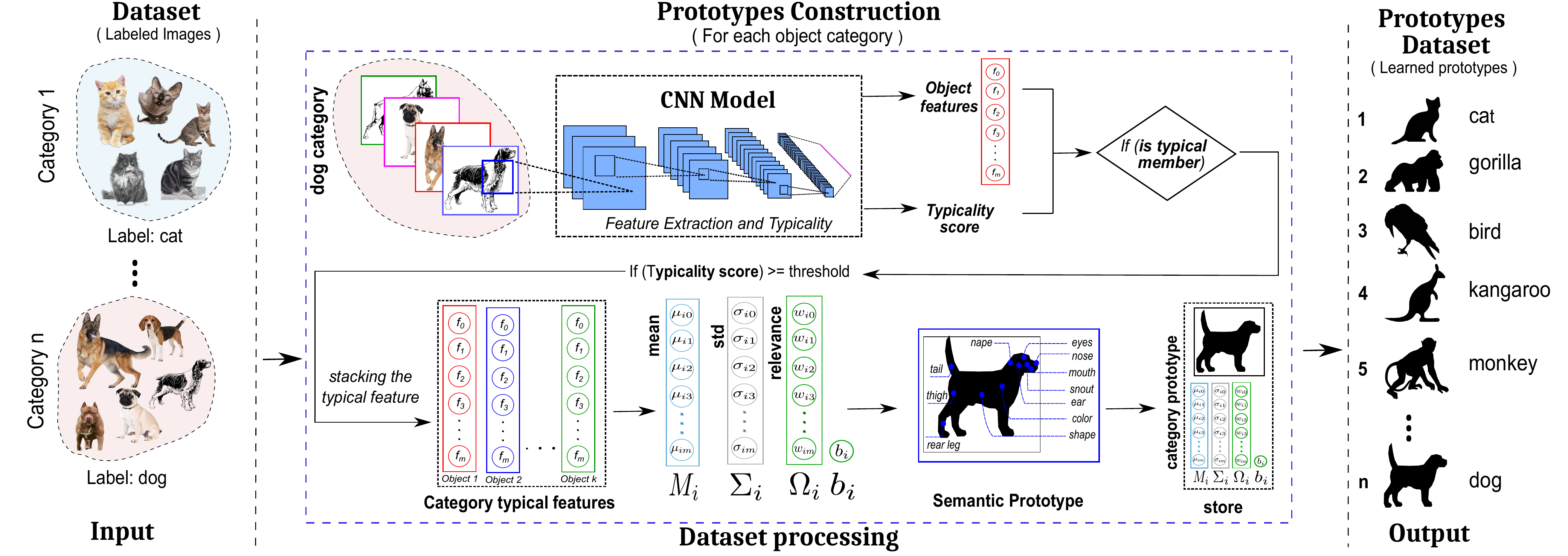}
	\end{center}
	    \vspace*{-\baselineskip}
	\caption{\textbf{Off-line construction of the semantic prototype dataset}. Given a labeled images dataset, we compute our semantic prototype representation for each object category present in the dataset. The diagram details the offline computation of a semantic prototype for a specific category.}
	\label{fig:prototype_construction} 
	
\end{figure*}

Since we have also considered the element's typicality within the category to compute our semantic prototype representation, the prototype construction requires image datasets of objects with annotations of the image typicality score. Considering that large image datasets~\cite{ILSVRC15, lin2014microsoft,lecun1998,krizhevsky2010convolutional} do not have annotations of image typicality, we use as typical objects of a category those elements that are classified as belonging to a specific category (those with $\sim100\%$ probability of membership).

Figure~\ref{fig:prototype_construction} shows the main steps and concepts of our prototype construction procedure.
Given a labeled dataset with images of objects, we extract the features and typicality score of objects' images for each object category. Next, for those object image features~(typical features) that have typicality scores higher than a threshold, we compute our semantic prototype representation (we use a threshold $\geq 98\%$). 
The resulting semantic prototypes dataset is used as prior knowledge in our procedures to introduce our CPM framework into deep learning models and evaluate it in real-world image processing tasks.

\section{\textbf{CPM model semantics in classification and description of object images}}
\label{sec:Tasks}

Rosch's experiments~\cite{rosch1973internal,rosch1975family,rosch1975cognitive,rosch1976structural,rosch1978principles} indicated that category prototypes are cognitive reference point in constructing concepts. 
We apply the Prototype Theory as a theoretical foundation to represent the semantic of the visual information lying in the basics components of a scene: objects. 
The observations on the Prototype Theory raise the following two questions: i) Can a model of perception system be developed in which objects are described using the same semantic features that are learned to identify and classify them? ii) How can the category's prototype be included as a reference point in the object global semantic description and classification tasks?

We address these two questions inspired by the human's approach to classifying and describing objects globally. Humans use the generalization and discrimination processes to build object descriptions highlighting their most distinctive features within the category. For instance, a typical human description: a dalmatian is a dog~(generalization ability to recognize the central semantic meaning of dog category) distinguished by its unique black or liver-colored spotted coat~(discrimination ability to detect the semantic distinctiveness of object within the dog category). 
Figure~\ref{fig:motivation} depicted our prototype-based classification and description  hypothesis, and Figure 5 illustrates our workflow  to model its main concepts.


\subsection{\textbf{Prototype-based Classification using CPM Model}}
\label{subsec:PSLayer}

\begin{figure*}[t!]
	\begin{center}
		\includegraphics[scale=0.5,width=0.95\linewidth]{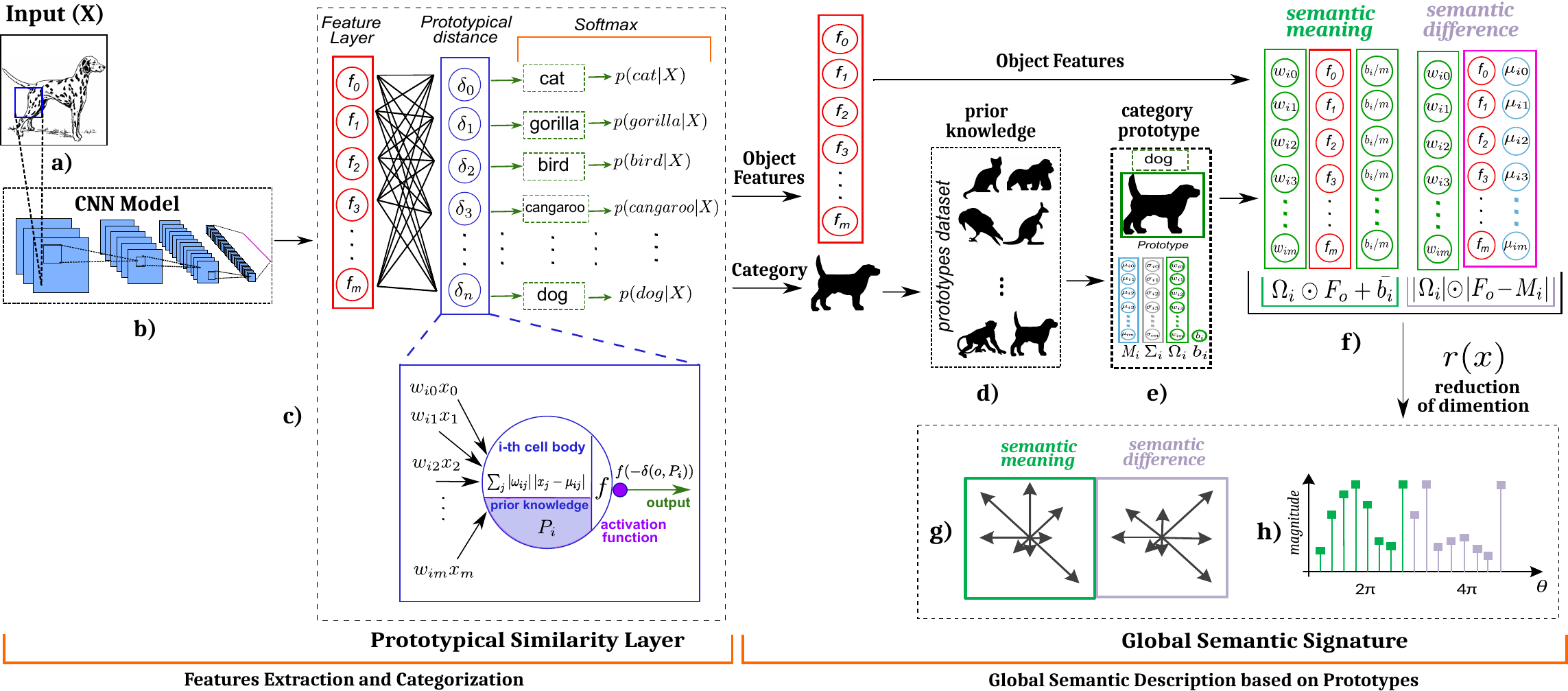}    
	\end{center}
	    \vspace*{-\baselineskip}
	\caption{\textbf{Prototype-based Models Workflow}. Our methodology comprises two main stages: \textit{\textbf{1)}} Feature extraction and Categorization based on prototypes and \textit{\textbf{2)}} transformation of CNN-object features into our Global Semantic Signature based on Prototypes.~\textit{a}) input image;~\textit{b)-c)} features extraction and classification using a pre-trained CNN classification model. Our Prototypical Similarity Layer (PS-Layer) is used to convert a common CNN-model into a prototype-based CNN-classification model;~\textit{d)} prototype dataset;~\textit{e)} category prototype selection;~\textit{f)} global semantic description of an object using category prototype;~\textit{g)} graphic representation of our Global Semantic Descriptor signature resulting from the dimensionality reduction function~($r(x)$); and~\textit{h)} Global Semantic Signature.}
	\label{fig:methodology}
\end{figure*}

This section introduces our CPM framework to simulate the prototype-based concept of categorization of Prototype Theory (Figure~\ref{fig:motivation} steps 1-3). To evaluate our framework in the image classification task, we propose a CNN-Layer that converts a common CNN classification model into a prototype-based classification. The diagram in Figure~\ref{fig:methodology}-c illustrates the internal structure of our new Prototypical Similarity Layer~(PS-Layer). The diagram shows the process of using the PS-Layer in a common CNN classification model. It highlights in purple the mathematical model of a PS-Layer neuron. Notice how the $i$-th neuron body keeps, as prior knowledge, the semantic prototype ($P_i$) of $i$-th category. The PS-Layer has many neurons as prototypes and categories~(see Figure~\ref{fig:methodology}-c) and uses as neuron output activation our prototypical distance to measure the object's semantic distinctiveness.

Similar to MPM model~\cite{zaki2003prototype,minda2002comparing}, the PS-Layer computes the probability with which an object $ o \in O$ is classified into i-$th$ category using the equation:
$ 
{P(c_i|o) = S(o,P_i)^{\gamma} / \sum_{k=1}^{n}S(o,P_k)^{\gamma}},
\label{eq:probability}
$
\noindent where $\gamma$ is the response-scaling parameter, and ${S(\emph{o},P_i) = \exp (-\alpha \times \delta(\emph{o},P_i))}$ is the similarity between object $ o \in O$ and the i-$th$ prototype $(P_i)$. Without loss of generality, and using the same MPM model assumptions~\cite{minda2002comparing,zaki2003prototype}, we set $\alpha$ and $\gamma$ to be equal to $1$. The classification probability of our PS-Layer can be rewritten as:
\begin{eqnarray} 
P(c_i|o) = \frac{\exp(-\delta(\emph{o},P_i))}{\sum_{k=1}^{n}\exp(-\delta(\emph{o},P_k))},
\label{eq:softmax_probability}
\end{eqnarray} 
where $\delta(\emph{o},P_i)$ is our prototypical distance. Note that our PS-Layer is a softmax function over the prototypical distance as probability distribution $ {P = \textit{softmax} \left(-\vec{\delta}(\emph{o},P_k)\right)}$. To simplify the PS-Layer neuron gradient computation, we added several constraints: \textit{i)} neuron weights must be non-negative, \Wlog, this allows to eliminate absolute value sign in the weights-term of our prototypical distance expression; \textit{ii)} L2-regularization is used to guarantee small weights values~\cite{zaki2003prototype}. Consequently, since $\mu_{ij} \in M_{i}$ is a constant, our PS-Layer neuron gradient:   
\begin{eqnarray}
\frac{\partial \delta}{\partial \omega} =
\begin{cases}
\mu_{i} - x, & \text{if } \omega x- \omega \mu_{i} \geq  0 \\
x - \mu_{i}, & \text{if } \omega x- \omega \mu_{i} < 0
\end{cases}
\label{eq:derivative_d}	
\end{eqnarray}

\noindent is as simple as common CNN neuron gradient ${\partial z}/ \partial \omega = x$. Then, the model that uses our PS-Layer can be trained using the same training conditions of a baseline CNN model.
Our prototype-based classification approach differs from other literature works~\cite{seo2003soft,wohlhart2013optimizing, jetley2015prototypical,snell2017prototypical} as follows:~i) the prototype  representation is based on the structural bases of typicality effects~\cite{rosch1975family, rosch1975cognitive,  rosch1976structural};
~ii) the similarity measure is also based on some psychological measures that assume the categories as heterogeneous semantic structures; 
~iii) regarding simplicity and scalability: our approach is less complex than other works in the literature. It is easy to use and convert a common CNN model into a prototype-based approach without making substantial changes to the original CNN-model architecture;~iv) regarding interpretability: 
 Since we are trying to capture the image membership degree, our PS-Layer provides greater interpretive power to CNN classification models due to simplicity and clear geometric interpretation of the object typicality concept.



\subsection{\textbf{Global Semantic Descriptor based on Prototypes}}
\label{subsec:GSDP descriptor}

\subsubsection{Semantic Meaning Vector}
Many cognitive neuroscience works have studied the effect of semantic meaning in object recognition task~\cite{tulving2007coding,martin2007representation,collins2013conceptual}. They observed that when an object is previously associated with semantic meaning in the brain, people are more prone to identify the object correctly. They also have shown that semantic associations allow a much faster recognition of an object, even when the task of object recognition becomes increasingly hard (varying points of view, occlusion). 
Moreover, the impressive performance of CNNs in object classification tasks fostered studies of possible links between CNN models and the visual system in the human brain. Cichy~\etal~\cite{cichy2017dynamics}, for instance, suggested that deep neural networks perform spatial arrangement representations like those performed by a human being. Khaligh~\etal~\cite{khaligh2014deep}, for their turn, concluded that the weighted combination of features in the last fully connected layer of CNNs could thoroughly explain the inferior temporal cortex in the human brain. We lay hold of these theoretical foundations to model our representation of the object's semantic meaning.



We redefine as semantic value of the object ${\emph{o} \in O}$ in the context of $i$-th category, the image score ${z= \sum_{m} \omega_{ij}f_j + b_{i}}$, where ${\omega_{ij} \in \Omega_{i}}$ is the relevance of $j$-th object feature. 
Note that the semantic value for an object is the value commonly used to object categorization in the softmax layer of CNN classification models. Hence, our descriptor approach based on prototypes assumes as the object semantic meaning vector, the  vector~(${\vec{z} = \Omega_{i}  \odot F_{o} + \vec{b_i}}$) constructed using the Hadamard product $\odot(\cdot)$ to compute the object semantic value. Our semantic meaning representation applies a bias vector to dissolve the bias value in each semantic vector component uniformly. Thus, we use the sum of each semantic meaning vector component to recovering the semantic value, \ie, $z = \sum_m \vec{z}$.



\subsubsection{Semantic Distinctiveness Vector}
\label{subsubsect:semantic_distance}

We stand for the semantic distinctiveness of an object for specific $c_i$-category as the semantic discrepancy between object features and features of the most prototypical element of $c_i$-category~($i$-th abstract prototype). 
Consequently, our approach assumes as semantic distinctiveness vector of an object, the semantic difference vector: ${\vec{\delta}_i = \left|\Omega_{i}\right| \odot \left|F_{o} - M_{i}\right|}$ that is constructed with the element-wise operations to compute the object prototypical distance. 
The semantic difference vector is the weighted residual vector $\Omega_{i}$ that is composed of the absolute different between each object feature and each feature of $i$-th category abstract prototype, \ie, $\left|F_{o} - M_{i}\right|$.
Therefore, similar to semantic meaning vector,  we use a sum of each semantic difference vector component to retrieve the object prototypical distance, \ie, $\delta(\emph{o},P_i)= \sum_m \vec{\delta}_i$.

\begin{algorithm} [t]
	\caption{Global Semantic Descriptor $\psi$}
	\label{alg:global_descriptor}
	\begin{algorithmic}[1]
		\BState \emph{\textbf{Input}}: Image of an object $ o $ 
		\BState \emph{\textbf{Output}}: GSDP signature~($\psi_o$)
		\BState \emph{\textbf{Prior-Data}}: Trained CNN-model $\Lambda$, $prototypes\_set$
		\State $ F_{o}, c_{i} \gets \Lambda.features\_and\_prediction(o)$
		\State $ {M_{i},\Sigma_{i},\Omega_{i},b_{i}} \gets prototypes\_set(c_i)$
		\State $ \textit{low-meaning} \gets r\left({F_o,\Omega_{i},b_{i},\textit{semantic meaning}}\right)$
		\State $ \textit{low-difference} \gets r\left({\left|F_{o} - M_{i}\right|,\Omega_{i},b_{i},\textit{distinctiveness}}\right)$
		\State \Return $ \textit{low-meaning} \oplus \textit{low-difference} $
	\end{algorithmic}
\end{algorithm}


Figure~\ref{fig:methodology} show an overview of our prototype-based description model. 
After the feature extraction and categorization processes (Figure~\ref{fig:methodology}a-c), we use the corresponding category prototype for describing the object features semantically. We show the steps that introduce the category prototype into the global semantic description of the object's features in Figure~\ref{fig:methodology}d-h). A drawback of this semantic representation~(Figure~\ref{fig:methodology}-f) is having high dimensionality since it is based on the semantic meaning vector~($\vec{z}$) and the semantic difference vector~($\vec{\delta}$). The large dimensionality of our feature vectors could make its use unfeasible in common computer vision tasks~\cite{han2017scnet,kim2017fcss}. Figure~\ref{fig:methodology} and Algorithm~\ref{alg:global_descriptor} detail the main steps of our approach; note that the steps follow the same workflow of human description hypotheses depicted in Figure~\ref{fig:motivation}.



\subsubsection{Dimensionality Reduction}

Discarding features, from the Prototypes Theory perspective, is not suitable when applied to the semantic space due to the absence of necessary and sufficient definitions to categorize an object (Table~\ref{table:prototypicality-effects}: intensional non-discreteness). Discarding features might lead to discarding discriminatory ability among category elements~\cite{geeraerts2010theories}, since some objects within the category do not have some category typical features. For example, flying is a typical feature of the bird category, but a penguin is a bird that does not fly.

\begin{algorithm}[t]
	\caption{Dimensionality Reduction $r(x)$}
	\label{alg:dim_reduction}
	\begin{algorithmic}[1]
	
		\BState \emph{\textbf{Input}}:
		\textit{m-dim vector}~$\alpha$, $\Omega_{i}, b_{i},$ \textit{type}, \textit{size-option}
		
		\BState \emph{\textbf{Output}}:  semantic signature
		\\ \textit{\footnotesize//~Bias m-dimensional vector initialization}
		\State $ \bar{b_i} \gets \frac{b_i}{m} \qquad \qquad$  {\footnotesize //~ $ \bar{b_i}$ $(b_{i} = \sum_{m} \bar{b_i})$}
		\\ \textit{\footnotesize//~Auxiliary matrix initialization}
		\State $r \gets dimension\_from(\textit{size-option})$
		\State $ \chi_{r\times r} \gets shape(r,r) \: $ {\footnotesize//~square auxiliary matrix} \\ 
		\textit{\footnotesize//~Computing angles matrix from auxiliary matrix~$\chi_{r\times r}$}.
		\State $\Theta_{r\times r} = angles\_from( \chi_{r\times r})$
		\\ \textit{\footnotesize//~Computing \textit{semantic-vector} using the Hadamard product $\odot$}.


		\\ \textit{\footnotesize//~Finding the optimal configuration $p, q$}.
		\State $p \equiv 0\ (\textrm{mod}\ r)$, $q \equiv 0\ (\textrm{mod}\ r)$ and $p \cdot q = m$
		\\ \textit{\footnotesize//~Reshape semantic-feature to $(p, q)$ dimension}.
	
		\State $  S_{p \times q} = $  $reshape\_to\_matrix_{p\times q}(\vec{S})$
		
		\State $ signature \gets \left [  \right ]$
		\\ \textit{\footnotesize//~Sliding the angles-matrix (kernel) across features-vector  $  S_{p \times q} $ }
		\For {\textbf{each} $\Theta_{r\times r}$ \textbf{in} $S_{p \times q} $}
		
		 \textit{\footnotesize//~Computing the semantic-gradient matrix}
		\State $G^{k} \gets vectors(\Theta_{r\times r},\left|S_{p\times q}^{k}\right|, sign(S_{p\times q}^{k}))$
		\State \textit{\footnotesize//~Computing 8D-histogram of gradients}
		\For {$l \in [1,...,8]; \theta_l = l  \cdot \frac{\pi}{4} $}
	    \State  \textit{\footnotesize//~Gradients are quantified with each angular bin}

		\State  $8D$-$hist^{k}[l]=\sum G^{k}(\theta), \forall \theta: \theta_l-45 < \theta \leq \theta_l$  
		\EndFor
		 \State  \textit{\footnotesize//~Adding 8d-histogram to final semantic signature}
		\State $ signature \gets signature \oplus 8D$-$hist^{k} $
		\EndFor
		\BState \Return $signature $    
	\end{algorithmic}
\end{algorithm}

Several dimensionality reduction algorithms such as PCA~\cite{abdi2010principal} and NMF~\cite{lee2001algorithms} are based on discarding features that do not generate a meaningful variation. Although these approaches work on some tasks, we can lose the ability of data interpretation after applying these algorithms~\cite{abdi2010principal}. In order to encode our representation with low dimensionality while encapsulating some interpretable properties of the object's image, we propose a simple transformation function~$r(x)$ to compress our global semantic representation of the object's features~(Figure~\ref{fig:methodology}-f) into a low dimensional signature~(Figure~\ref{fig:methodology}-h).

We propose the transformation $r(x)$ to reduce the dimensionality of our image semantic representation~(Figure~\ref{fig:methodology}-f) while retaining, in the final descriptor, properties such as object semantic meaning and object semantic difference~(typicality vector). 
Our final descriptor~$\psi$ is computed by concatenating the corresponding signatures of semantic meaning vector~$\vec{z}$ and semantic difference vector~$\vec{\delta}$ compressed with the transformation $r(x)$.
Algorithm~\ref{alg:dim_reduction} describes the steps of our dimensionality transformation function. The workflow can be summarized in nine main steps: \textbf{1)} Transform the learned bias value $ b_{i}$ in the $ m $-dimensional vector $\bar{b_i} \in \mathbb{R}^{m}$, $\bar{b_{ij}}$ = $b_{i}/m$ such that ${b_i = \sum_m \bar{b_i}}$;\textbf{2)} Compute the auxiliary matrix $ \chi_{r\times r}$ based on the descriptor signature size desired~(size-option parameter); \textbf{3)} Compute the angles matrix $\Theta_{r\times r}$ using the angles formed by the position of each auxiliary matrix cell with respect to auxiliary matrix $\chi_{r\times r}$ center. To achieve uniqueness the diagonal angles are evenly distributed between among its neighboring angles magnitudes (max and min angles); \textbf{4)} Compute our high-dimensional semantic representation based on prototypes (Figure~\ref{fig:methodology}-f); \textbf{5)} Resize the semantic representation in the best 2D dimensional matrix configuration~($ p \times q $) whose dimensions are multiples of $r$~(auxiliary matrix dimension);
~\textbf{6-7)}~Slide the angles-matrix (as kernel) across features-matrix $S_{p\times q}$ and create an unitary semantic gradient for each angles-matrix mapped within the features-matrix. Each semantic gradient is constructed using the angle matrix $\Theta_{r\times r}$, magnitude and sign of features-matrix values; ~\textbf{8)} Reduce each semantic gradient $G^{k}$ to a 8D-histogram similarly to SIFT~\cite{lowe2004SIFT};~\textbf{9)} Concatenate, for each semantic gradient, the corresponding 8D-histogram resulted of flow 6--8.

Note that, similar to our global semantic representation of object features (Figure~\ref{fig:methodology}-f), the GSDP-signature holds important properties: the first half of our semantic signature preserves the objects semantic meaning, \ie, ${\sum_{l=0}^{\left | \psi  \right |/2} \psi[l] = z}$; the second half retains the object semantic difference(object typicality), formally, ${\sum_{l=\left|\psi  \right |/2}^{\left | \psi \right |} \psi[l] = \delta(\emph{o},P_i)}$. 
Additionally, our descriptor can construct semantic representations for: \textit{i)} an object, and  \textit{ii)} an abstract prototype (ideal category member). 

\section{\textbf{Experiments and Results}}
To verify that our Prototype Framework contributes to the success of semantic extraction, capturing the semantic core of the category, simulating the visual representation degree of each image as well as the usability of that semantic information in image processing tasks, we performed the following experiments.
First, we qualitatively analyze the semantics captured by our CPM model. Second, we evaluate the performance and usability of our CPM model in the image classification task. Third, to validate the assumption that the core semantic information of a category must be invariant to the training dataset, we carry out cross-dataset transfer-learning experiments based solely on the information captured by our prototype representation. Finally, since our GSDP descriptor is based on the object typicality concept,  we evaluate the performance of our semantic description approach in image clustering and classification tasks.

\paragraph{Datasets} We conducted our experiments on five image datasets.
~The off-line prototype computation process and the CPM model representation were conducted using MNIST~\cite{lecun1998}, CIFAR10, CIFAR100~\cite{krizhevsky2010convolutional}, and ImageNet\cite{ILSVRC15} datasets.
PS-Layer classification performance was evaluated using MNIST~\cite{lecun1998}, CIFAR10, and CIFAR100~\cite{krizhevsky2010convolutional}.
We evaluated the prototype-based transfer learning performance and our GSDP descriptor performance using the ImageNet~\cite{ILSVRC15} and Coco~\cite{lin2014microsoft} as real images datasets.

\paragraph{Backbone Networks} We evaluated our CPM representation using CNN architectures based on LeNet~\cite{lecun1998} and Deep Belief Network~\cite{krizhevsky2010convolutional}  for MNIST and CIFAR datasets, respectively.  We assessed the PS-Layer performance with the following networks: sMNIST~\cite{lecun1998}, sCF10~\cite{krizhevsky2010convolutional}, sCF100~\cite{krizhevsky2010convolutional}, vggCF10~\cite{liu2015very}, and vggC100~\cite{liu2015very}. We also conducted cross-dataset experiments in ImageNet~\cite{ILSVRC15} and Coco~\cite{lin2014microsoft} using VGG16~\cite{simonyan2014very} and ResNet50~\cite{he2016deep} as backbone networks of our GSDP  representation and prototype-based transfer learning experiments.

\subsection{\textbf{Computational Prototype Model}}

Due to the lack of a well-defined metric to quantify whether a framework correctly captures the semantic meaning of a category and annotated images with the object typicality score, we used four assessment approaches to analyze the semantics underlying our CPM model, namely: Semantic prototype encoding, Central and Peripheral meaning, Prototypical Organization, and Image Typicality Score.

\subsubsection{Semantic prototype encoding}
\begin{figure}[t]
	\begin{center}
		\includegraphics[width=0.8\linewidth]{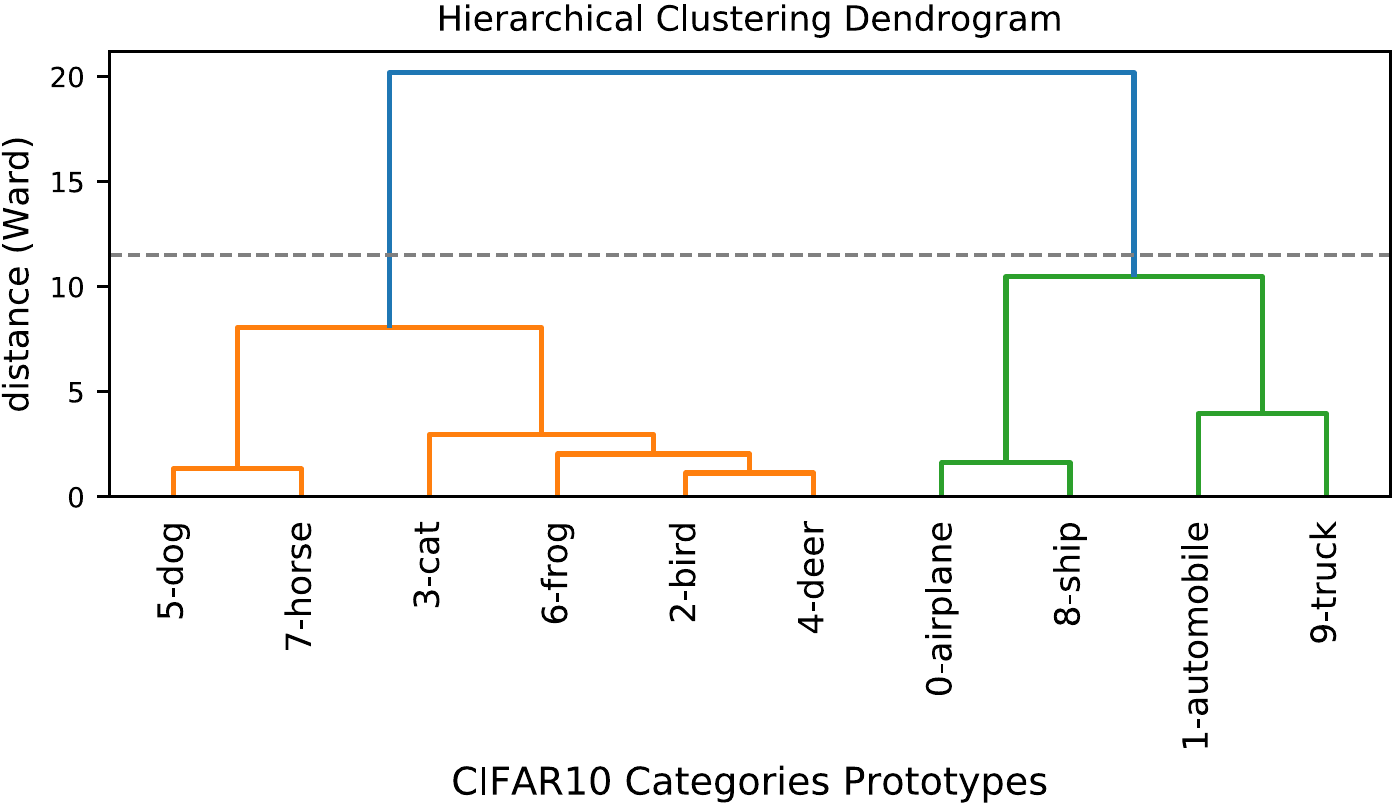}
	\end{center}
	\vspace*{-\baselineskip}
	\caption{\textbf{Hierarchical clustering of CIFAR10 semantic prototypes}.}
	\label{fig:cifar10_prototypes}
\end{figure}

We analyzed the semantics behind our semantic prototype representation by conducting the hierarchical clustering of our categories' semantic prototypes, which illustrates the hierarchical semantic organization of a specific image dataset. Figure~\ref{fig:cifar10_prototypes} shows an example of a dendrogram obtained when using the semantic prototypes computed in CIFAR10. Notice that our semantic categories' representations distribute the CIFAR10-dataset, achieving a hierarchical semantic organization. For example, two macro-categories are visible in Figure~\ref{fig:cifar10_prototypes}: animals and transport vehicles. It is noteworthy that this last macro-category is also semantically interpreted by our representation as non-ground vehicles and ground vehicles.

\subsubsection{Central and Peripheral meaning}

\begin{figure}[t]
	\begin{center}
		\includegraphics[width=0.98\linewidth]{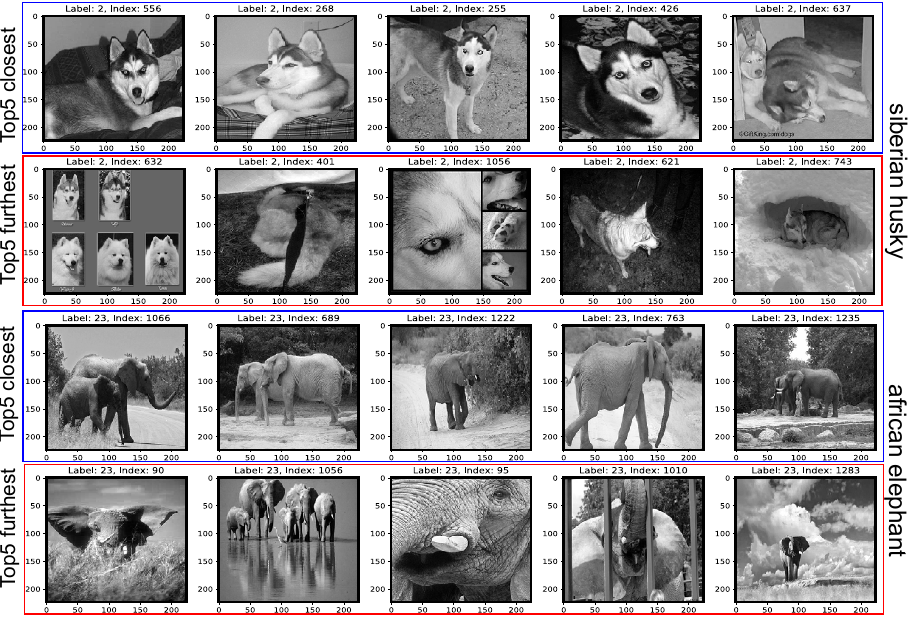}
	\end{center}
	\vspace*{-\baselineskip}
	\caption{\textbf{Central and Peripheral meaning captured by our CPM model.}
	From left to right:  Top-$5$ elements closest (in blue) to the semantic prototype of the corresponding category; and Top-$5$ elements furthest (red) from the category semantic prototype. Index value represents the image position within the category dataset. 
	}
	\label{fig:top5_ImageNet}    

\end{figure}

We observed the visual representativeness of those elements allocated by our CPM model in the center and periphery of the category. We aim to study the visual representativeness, \ie, typicality, of category members closest and farthest from the category semantic center (our abstract prototype). We extract features from an object's image with a CNN model and compute the prototypical distance for all $ i $-th category members. Next, the objects' images are ranked in ascending order based on their prototypical distance score. Figure~\ref{fig:top5_ImageNet} shows examples of central~(Top-$5$ closest) and peripheral~(Top-$5$ farthest) meaning captured by our CPM model in ImageNet categories using VGG16-model as an image feature extractor.  Note that our proposal finds typical elements (Top-$5$ closest) images with distinctive features in the category. Members with fewer characteristic features, or little readable, are placed in the periphery (Top-$5$ farthest) away from the category central semantic meaning. However, they are keeping the category features since they still belong to the category. 


\begin{figure}[t]
	\begin{center}
		\includegraphics[scale=0.8,width=0.9\linewidth]{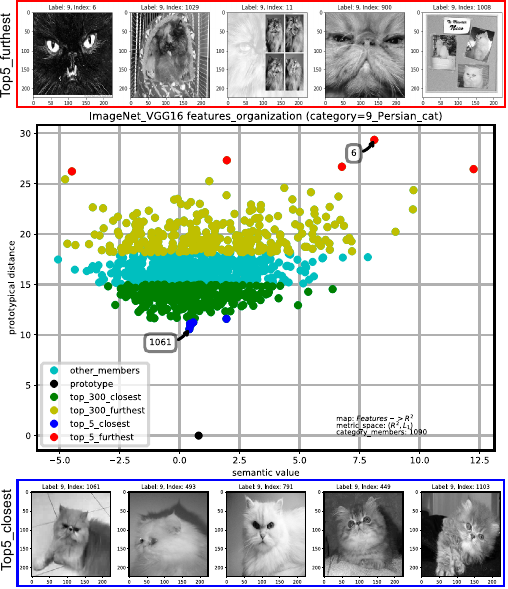}
		\caption{\textbf{Prototypical organization within categories.} The internal structure of the \textit{Persian cat} category in ImageNet. Each category member is represented with its VGG16 image features. We represented with color degrees the category's internal disposition respect its prototype. In the bottom and on the top, from left to right, the mapped Top-$5$ elements closest (in blue) and furthest (in red) to the mapped semantic prototype (in black). The image dataset index of the first Top-$5$ element is annotated inside the black box.}
		\label{fig:prototypical_organization}
	\end{center}
	\vspace*{-\baselineskip}
\end{figure}

\subsubsection{Prototypical Organization} 
We conducted experiments to analyze the internal semantic structure of the category applying the CPM model constraints. Visualizing the category's internal structure is infeasible in $ m $-dimensional features space. Since most data visualization methods are based on feature discarding, we used topology techniques to perform some continuous deformations of the object features and preserve some object semantic properties. The proposed map allows making an object's image interpretation based on all observed features.


Our function ${\rho: {F}_{c_i} \to \mathbb{R}^{2}\mid \rho(F_o) = p(z_o,\delta(o,P_i))}$ maps object's image features from  $({F}_{c_i},\delta)$ metric space to $(\mathbb{R}^{2},L1)$ metric space ($L_1$ is L1-norm condition) using its semantic value and its prototypical distance.  Thus, using our CPM constraints we can show that:~${\delta(o_1,o_2) \leq L_1(p_1,p_2) \leq 2 \delta(o_1,o_2)}$, \ie,~$\rho $ is continuous, which means that every element of  $\rho(o_1)$ neighborhood in $(\mathbb{R}^{2},L_1)$ also belongs into $o_1$ neighborhood  in $({F}_{c_i},\delta)$. Consequently, the observed behavior of $i$-th category internal structure -- in terms of distance metrics -- in $(\mathbb{R}^{2},L_1)$  is equivalent to the behavior in feature metric space $({F}_{c_i},\delta)$. Figure~\ref{fig:prototypical_organization} shows an example of the internal semantic structure captured by our CPM-model. 
Our experiments showed that the semantic value and the prototypical distance place the object in a unique semantic position within the category's internal structure. 
Notice that the internal structure of the category shows a prototypical organization of its elements in $(\mathbb{R}^{2},L_1)$ metric space.



\begin{figure} 
	\begin{center}
		\includegraphics[scale=0.98,width=0.95\linewidth]{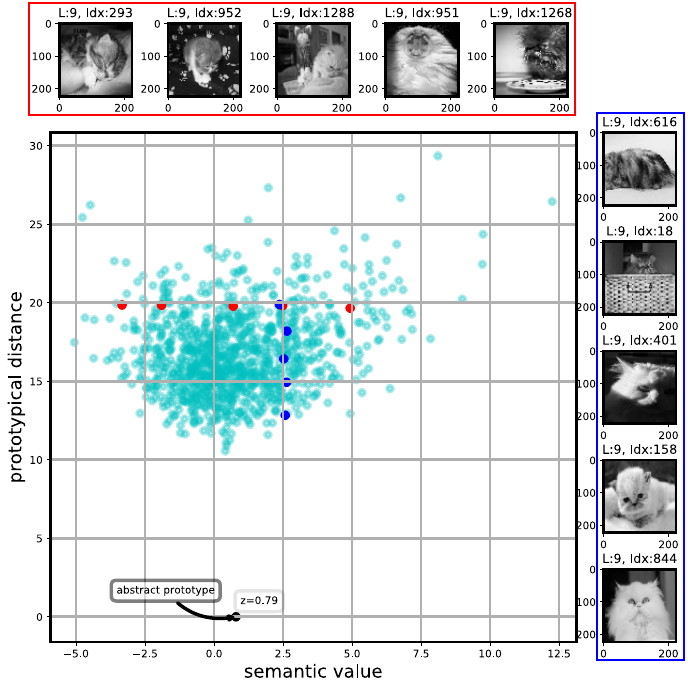}
		   \vspace*{-\baselineskip}
		\caption{\textbf{Typicality score analysis}. Images with the same prototypical distance and different semantic values (red) have similar representativeness within the category, and category members with different prototypical distance and same semantic value~(in blue) are visually different. We also observe that the image visual representativeness~(typicality) decreases as prototypical distance increases. The VGG16 model was used to extract image features.}
		\label{fig:semanticV_vs_prototypicalD}
	\end{center}
	    \vspace*{-\baselineskip}
\end{figure}

\subsubsection{Image Typicality Score}
Our approach to visualize the internal structure of a category also allows observing other semantic phenomena related to the visual representativeness degree of an object's image. We conducted qualitative experiments to investigate the influence of variations of semantic value and prototypical distance on the image's visual representativeness. The experiments showed a small strength of a linear association between those two variables~(Pearson coefficient values between $-0.3$ and $0.3$). Figure~\ref{fig:semanticV_vs_prototypicalD} shows an example of our experiment.
 
Lake~\etal~\cite{lake2015deep} shows that the semantic value can be used as an indicator of the typicality of an input image.   In contrast to Lake~\etal's~results, our experiments in the ImageNet dataset with VGG16 and ResNet50 models showed that using the semantic value as typicality score of the object's image can be problematic. Mainly because objects with the same semantic value do not imply the same image typicality~(\eg,~images highlighted in blue in Figure~\ref{fig:semanticV_vs_prototypicalD}). Selecting the semantic value as image typicality score shares the same serious problems that CNNs still suffer from. Adding small noises or making small changes to the initial samples generates different predictions for these samples with high confidence~(adversarial samples~\cite{szegedy2014intriguing}), thus generating drastically different semantic values for very similar images. 
Our experiments did not allow us to generalize a behavior pattern between semantic value and image typicality score. 
However, the experiments carried out suggest that our prototypical distance can capture the representativeness degree of the object's image. We observe that as the prototypical distance increases,
~the visual typicality of the object's image decreases.


\subsection{\textbf{Prototype-based Classification}}

We used our PS-Layer to assess the performance of our CPM framework in classification tasks. The experiments were performed using baseline models the following CNN architectures: sMNIST~\cite{lecun1998}, sCF10~\cite{krizhevsky2010convolutional}, sCF100~\cite{krizhevsky2010convolutional}, vggCF10~\cite{liu2015very}, and vggC100~\cite{liu2015very}. The CNN-baseline models used differ in output categories number, model architecture, model depth, accuracy, and dataset size to evaluate our approach in different environments. For each CNN-baseline model, we replaced the softmax layer with our PS-Layer. We trained the resulting PS-Layer models using the same training conditions of its baseline CNN-model~(\ie, batch size, epochs, without data-augmentation, etc.). We evaluated several versions of PS-Layer models, changing the weights initialization method and version of our semantic distance function used as a prototypical similarity.
For each weights initialization method: from scratch, freezing, and pre-train, we used two versions of our semantic distance function inside the PS-Layer:~a) prototypical distance; and b) penalized prototypical distance. We penalized peripheral elements using our semantic edges constraints (see Equation~\ref{eq:semantic_edges}). Consequently, for each baseline CNN model, we evaluated six PS-Layer model versions:fromscratch-a, fromscratch-b, freezing-a, freezing-b, pretrain-a, pretrain-b. Notice that unlike other prototype learning approaches~\cite{Ma2013,Ojeda2013,wohlhart2013optimizing, saleh2013object,Zhao2015,  jetley2015prototypical,saleh2016incorporating, Oyedotun2017, snell2017prototypical, drumond2017using, dong2018few, fort2018gaussian, Yang2018, allen2019infinite,angelov2020towards,  xiao2020tdapnet, garnot2021leveraging}, our prototype representation is not updated during the training process. Because the lack of annotated data with the typicality information prevents end-to-end training, our main goal is to evaluate the performance and robustness of the semantic information encapsulated in our prototype's representation.

\begin{table}[t]
	\centering
	\caption{Accuracy achieved by sCF10 versions using our PS-Layer in CIFAR10 dataset (best in bold).}
    \label{table:sCIFAR10_performance}	
	\resizebox{\columnwidth}{!}{%
	\begin{tabular}{|l|lc|cl|c|c|}
		\hline
		\multicolumn{1}{|c|}{\multirow{3}{*}{Model}} & \multicolumn{4}{c|}{Test}                                                              & \multicolumn{2}{c|}{Train}                            \\ \cline{2-7} 
		\multicolumn{1}{|c|}{}                       & \multicolumn{2}{c|}{Top1}                            & \multicolumn{2}{c|}{Top5}       & Top1                      & Top5                      \\ \cline{2-7} 
		\multicolumn{1}{|c|}{}                       & \multicolumn{1}{c|}{Mean$\pm$Std} & \multicolumn{1}{c|}{Max} & \multicolumn{1}{c|}{Mean$\pm$Std} & Max & \multicolumn{1}{c|}{Mean$\pm$Std} & \multicolumn{1}{c|}{Mean$\pm$Std} \\ 
		\hline \hline
		sCF10\cite{krizhevsky2010convolutional} & 69.53$\pm$2.18& 72.11 & 97.43$\pm$.42 & 98.09 & 74.32$\pm$2.74 & 98.23$\pm$.41                      \\ \hline \hline
		fromscratch-a &  73.42$\pm$.47& 74.05& 98.0$\pm$.26&98.23& 79.57$\pm$.76 & 98.95$\pm$.14  \\
		fromscratch-b & 73.71$\pm$.71& 74.96&98.01$\pm$.40 & 98.75   & 79.49$\pm$.81& 98.90$\pm$.20   \\
		freezing-a & 64.54$\pm$.13& 64.80&96.56$\pm$.04& 96.64   & 68.53$\pm$.09 & 97.26$\pm$.05\\
		freezing-b & 67.44$\pm$.08 & 67.54 & 97.29$\pm$.03 & 97.33 & 71.37$\pm$.06 & 98.05$\pm$.04\\
		pretrain-a & 75.84$\pm$.62& \textbf{76.84}& \textbf{98.45$\pm$.14}       & \textbf{98.64}    & \textbf{82.54$\pm$1.06}& \textbf{99.23$\pm$.11} \\
		pretrain-b &\textbf{75.87$\pm$.42}& \multicolumn{1}{c|}{76.47} & \multicolumn{1}{c}{98.30$\pm$.14}      & 98.55    & \multicolumn{1}{c|}{82.25$\pm$.55}     & \multicolumn{1}{c|}{99.20$\pm$.07}     \\ \hline
	\end{tabular}
    }
\end{table}

\begin{figure}[t]
	\centering
	\includegraphics[width=0.98\linewidth]{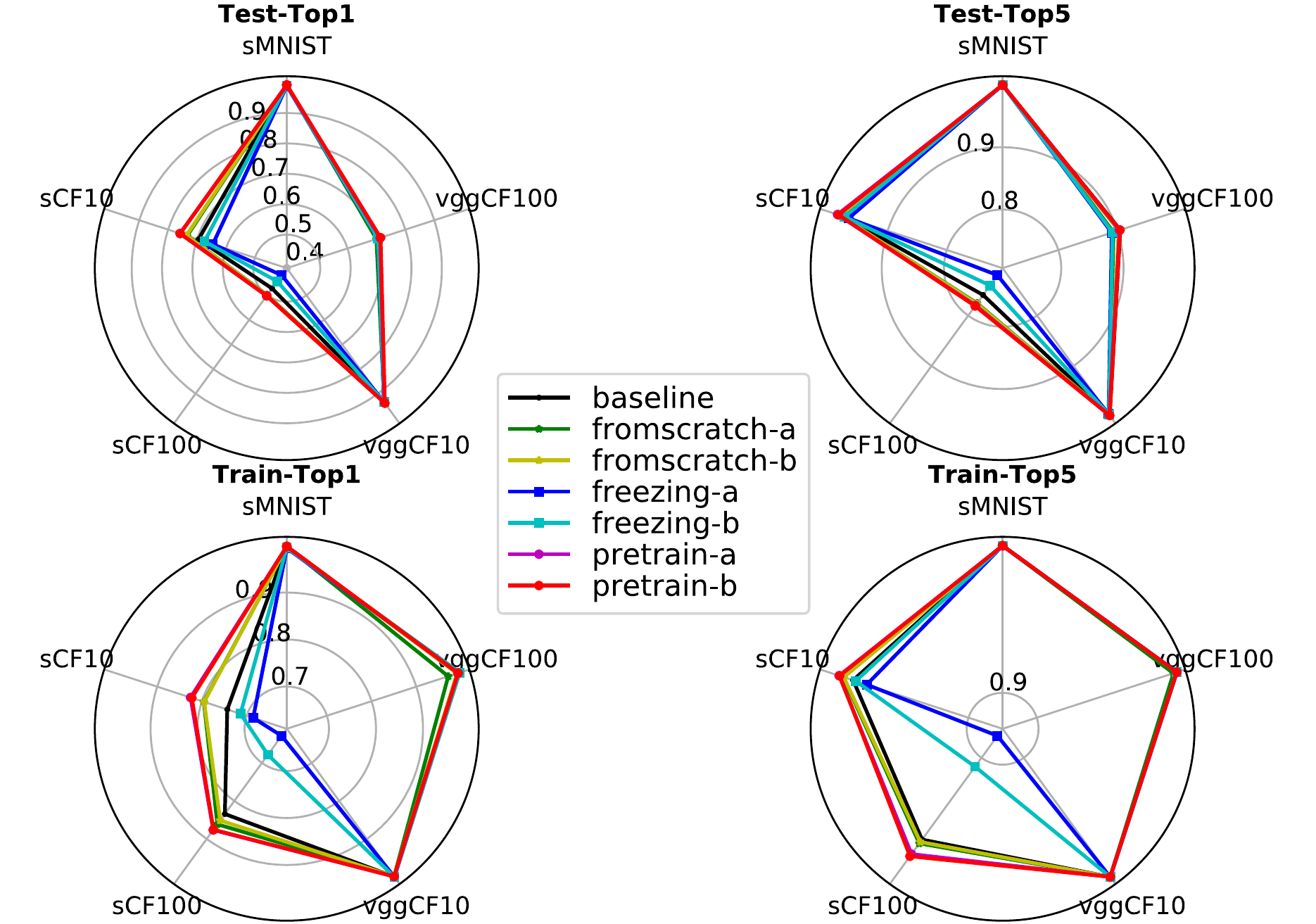} 
	\caption{\textbf{PS-Layer performance summary}. Classification accuracy overview of each baseline CNN model \textit{versus} our PS-Layer versions. Each circle summarizes the metrics performance (\textit{Test}-Top-$1$, \textit{Test}-Top-$5$, \textit{Train}-Top-$1$, \textit{Train}-Top-$5$) of each case study analyzed. Accuracy values were normalized between~$[0-1]$.}
	\label{fig:PS_layer_accuracy}
  	\vspace*{-\baselineskip}
\end{figure}

Table~\ref{table:sCIFAR10_performance} shows the performance for each PS-Layer version based on sCF10 model architecture~(baseline). The baseline CNN-model~(without PS-Layer) is in the first table row, separated from other PS-Layer models. Mean and Std accuracy values were computed using $10$ trained instances of each model version. 
Figure~\ref{fig:PS_layer_accuracy} summarizes the performance of each PS-Layer model version for each CNN-baseline model used as a case study. The experimental results show that the PS-Layer pre-train versions~(highlighted in magenta and red) outperform the baseline CNN-model (black) in all architectures analyzed. 
It is noteworthy that our PS-Layer can achieve good performance while provides greater interpretative power to CNN models.


\subsection{\textbf{Prototype-based Transfer-Learning}}

\begin{figure}[t]
	\centering
	\includegraphics[width=0.98\linewidth]{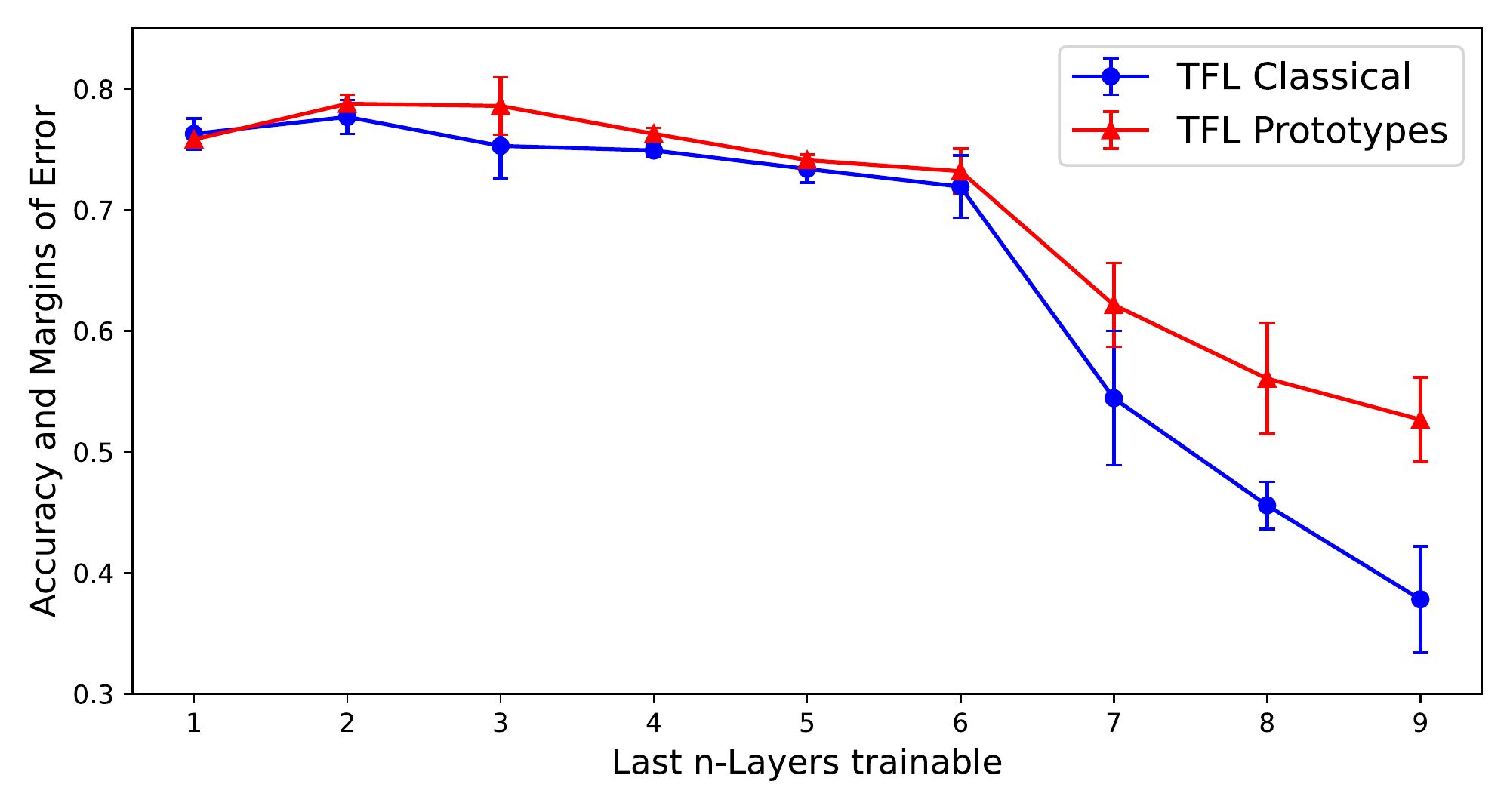} 
		\vspace*{-\baselineskip}
	\caption{\textbf{Prototype-based transfer-learning summary}. Classification accuracy overview of VGG16-based model \textit{versus} our VGG16-based PS-Layer model in cross-dataset transfer learning. 
	}
	\label{fig:PS_layer_tfl}
\end{figure}

Assuming that the central meaning of a learned category is invariant to the appearance of atypical members~\cite{rosch1975family,rosch1975cognitive}; we also evaluated on transfer learning (TFL) task the robustness of the semantic information encapsulated in our prototype representation. Classic transfer learning involves taking a pre-trained neural network and adapts the neural network to a new, different data set.  In contrast, we performed transfer learning experiments using only the semantic information encapsulated in pre-computed prototypes. Similar to the previous experiments, we used our PS-Layer to compare the classical transfer learning approach versus a prototype-based knowledge transfer learning approach.

Using our category prototyping approach (see Figure~\ref{fig:prototype_construction}),  we computed category prototypes in the ImageNet dataset (see details in supplementary material). We used them as prior knowledge to classify images in the COCO dataset. 
The TFL classical approach was evaluated using a CNN model based on VGG16 as a backbone network. The prototype-based TFL approach used VGG16 architecture plus PS-Layer with network weights initialized randomly. We accomplished a performance analysis of both approaches, training (under the same conditions) different model instances and changing (increasing) some parameters such as the number of trainable/frozen layers. These experiments can be understood as a simple ablation study to observe how performance degrades in both approaches as model components degrade.
Figure~\ref{fig:PS_layer_tfl} summarises the results of each experiment performed. 


\subsection{\textbf{Prototype-based image Global Descriptor}}


\begin{figure}[b]
	\centering
	\subfigure[]
	{
		\includegraphics[width=0.95\linewidth]{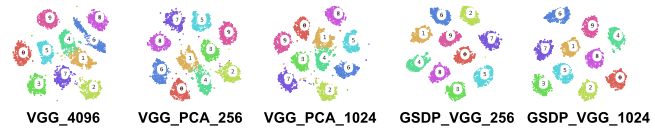}
		\label{fig:t-SNEvgg16} 
	}
	\vspace*{-\baselineskip}
	\subfigure[]
	{
		\includegraphics[width=0.95\linewidth]{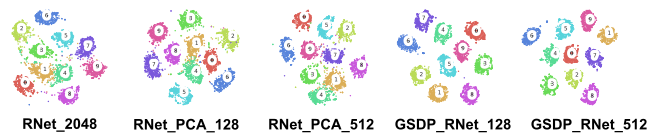}
		\label{fig:t-SNEresnet} 
	}
	\caption
	{
		\textbf{t-SNE visualization.} 
	 t-SNE visualizations of first $10$ categories of ImageNet dataset using features constructed with \textit{a)} VGG16 and \textit{b)} ResNet50 models. Feature length is shown in the corresponding caption.
	}
	\label{fig:t-SNE}
	\vspace*{-\baselineskip}
\end{figure}

\subsubsection{Signature Information Analysis}  
\label{subsection:signature_information}

Our GSDP descriptor uses category prototypes as a semantic distinctiveness generator of signatures for category members. Elements with similar semantic meanings and share similar semantic differences with the abstract prototype will have similar GSDP semantic signatures~(family resemblance concept). In other words, the abstract prototype can be interpreted as a DNA chain that stands for the typical features of category members. 
Since the t-SNE algorithm~\cite{maaten2008visualizing} can preserve the local structure, we used it to analyze the element neighborhood in $ m $-dimensional embedded space. 
We analyzed the discriminative power and t-SNE visualization performance of our GSDP semantic representation \textit{versus} features extracted using CNN models. 
We performed the t-SNE visualization experiment for features-family constituted by CNN features, corresponding GSDP semantic signatures, and reduced PCA versions of CNN-features (we reduced CNN-features to the same GSDP feature dimensions). Figure~\ref{fig:t-SNE} shows an example of t-SNE algorithm performance with VGG16 and Resnet features-family using Euclidean distance as similarity measure and $50$ as perplexity value. Note how GSDP representations achieved the best performance on each feature family.

\subsubsection{Performance Evaluation}
We evaluated our image semantic encoding performance in image clustering applications. 
According to observations of Yang~\etal~\cite{yang2016joint}, image features representations can generalize well when transferred to other tasks if they achieved good performance in the image clustering task. Based on these observations, we evaluated the GSDP descriptor performance in the clustering task 
by comparing its K-Means clustering metrics in ImageNet and Coco datasets. We compared our representation in the ImageNet dataset against:1) traditional handcraft image global descriptors: GIST~\cite{oliva2001modeling}, LBP~\cite{ojala2002multiresolution}, HOG~\cite{dalal2005histograms}, Color64~\cite{li2007texture}, Color\_Hist~\cite{song2004content}, Hu\_H\_CH~\cite{haralick1973textural,hu1962visual,song2004content}; 2) deep learning images features trained on ImageNet: VGG16 features and  ResNet50 features (and PCA-reduced versions). 



\begin{table}[t!]
	\caption{K-Means cluster metrics achieved for each evaluated global image representation in the first $20$ categories of ImageNet and Coco datasets (best in bold). 
	}
	\centering
	\label{table:Kmeans_metrics}
	\resizebox{\columnwidth}{!}{
	\begin{tabular}{|l|c|c|ccccc|} 
		\hline
		\multicolumn{1}{|c|}{\multirow{2}{*}{\textbf{Descriptor} }} & \multirow{2}{*}{\textbf{Size} } & \multirow{2}{*}{\textbf{FPS} } & \multicolumn{5}{c|}{\textbf{Metrics Scores} }                                                                                          \\ 
		\cline{4-8}
		\multicolumn{1}{|c|}{}                                      &                                 &                                & \multicolumn{1}{c|}{H}   & \multicolumn{1}{c|}{C}   & \multicolumn{1}{c|}{V}   & \multicolumn{1}{c|}{ARI} & AMI                        \\ 
		\hhline{|========|}
		\multicolumn{8}{|c|}{Handcraft Features Performance on ImageNet\cite{ILSVRC15}}                                                                                                                                                                                                            \\ 
		\hline
		GIST~\cite{oliva2001modeling}                                                       & \multicolumn{1}{l|}{960}        & \multicolumn{1}{l|}{0.82}      & \multicolumn{1}{l}{0.05} & \multicolumn{1}{l}{0.05} & \multicolumn{1}{l}{0.05} & \multicolumn{1}{l}{0.01} & \multicolumn{1}{l|}{0.05}  \\
		LBP~\cite{ojala2002multiresolution}                                                         & 512                             & 0.72                           & 0.02                     & 0.03                     & 0.03                     & 0.01                     & 0.02                       \\
		HOG~\cite{dalal2005histograms}                                                         & 1960                            & 33                             & 0.04                     & 0.04                     & 0.04                     & 0.01                     & 0.03                       \\
		Color64~\cite{li2007texture}                                                     & 64                              & 8                              & 0.12                     & 0.12                     & 0.12                     & 0.04                     & 0.11                       \\
		Color\_Hist\cite{song2004content}                                                 & 512                             & 26                             & 0.08                     & 0.08                     & 0.08                     & 0.03                     & 0.07                       \\
		Hu\_H\_CH~\cite{haralick1973textural,hu1962visual,song2004content}                                                     & 532                             & 6.9                            & 0.04                     & 0.04                     & 0.04                     & 0.01                     & 0.02                       \\ 
		\hhline{|========|}
		\multicolumn{8}{|c|}{Deep Features Performance on ImageNet\cite{ILSVRC15}}                                                                                                                                                                                                               \\ 
		\hline
		VGG16~\cite{simonyan2014very}                                                      & 4096                            & 15                             & 0,87                     & 0,88                     & 0,88                     & 0,78                     & 0,87                       \\
		VGG\_PCA\_256                                               & 256                             & 12.5                           & 0,89                     & 0,90                     & 0,89                     & 0,82                     & 0,89                       \\
		VGG\_PCA\_1024                                              & 1024                            & 12.5                           & 0,89                     & 0,89                     & 0,89                     & 0,81                     & 0,89                       \\
		GSDP\_VGG\_256~(our)                                              & 256                             & 12.8                           & \textbf{0,97}            & \textbf{0,99}            & \textbf{0,98}            & \textbf{0,93}            & \textbf{0,97}              \\
		GSDP\_VGG\_1024~(our)                                             & 1024                            & 11.6                           & 0,94                     & 0,98                     & 0,96                     & 0,84                     & 0,94                       \\ 
		\hline
		ResNet50~\cite{he2016deep}                                                    & 2048                            & 10.6                           & 0,88                     & 0,90                     & 0,89                     & 0,78                     & 0,88                       \\
		ResNet50\_PCA\_128                                              & 128                             & 12.5                           & 0,88                     & 0,88                     & 0,88                     & 0,81                     & 0,88                       \\
		ResNet50\_PCA\_512                                              & 512                             & 12.5                           & 0,89                     & 0,90                     & 0,90                     & 0,82                     & 0,89                       \\
		GSDP\_RNet\_128~(our)                                             & 128                             & 9.6                            & \textbf{ 0,97}           & \textbf{0,98}            & \textbf{0,98}            & \textbf{0,93}            & \textbf{0,97}              \\
		GSDP\_RNet\_512~(our)                                             & 512                             & 9                              & 0,91                     & 0,97                     & 0,94                     & 0,73                     & 0,91                       \\ 
		\hhline{|========|}
		\multicolumn{8}{|c|}{Deep Features Performance on Coco\cite{lin2014microsoft}(CrossDataset)}                                                                                                                                                                                                     \\ 
		\hline
		VGG16~\cite{simonyan2014very}
		& 4096          &15         & \multicolumn{1}{l}{0.32}     & \multicolumn{1}{l}{0.34}     & \multicolumn{1}{l}{0.33}     & \multicolumn{1}{l}{0.15}     & \multicolumn{1}{l|}{0.31}      \\
		VGG\_PCA\_256
		&256         & 12.5         & \multicolumn{1}{l}{0.35}     & \multicolumn{1}{l}{0.37}     & \multicolumn{1}{l}{0.36}     & \multicolumn{1}{l}{0.19}     & \multicolumn{1}{l|}{0.34}      \\
		VGG\_PCA\_1024
		& 1024           & 12.5         & \multicolumn{1}{l}{0.35}     & \multicolumn{1}{l}{0.37}     & \multicolumn{1}{l}{0.36}     & \multicolumn{1}{l}{0.18}     & \multicolumn{1}{l|}{0.34}      \\
		GSDP\_VGG\_256~(our)
		& 256          & 12.8        & \multicolumn{1}{l}{\textbf{0.47}}     & \multicolumn{1}{l}{\textbf{0.72}}     & \multicolumn{1}{l}{\textbf{0.57}}     & \multicolumn{1}{l}{\textbf{0.23}}     & \multicolumn{1}{l|}{\textbf{0.56}}      \\
		GSDP\_VGG\_1024~(our)
		& 1024          & 11.6        & \multicolumn{1}{l}{0.46}     & \multicolumn{1}{l}{0.54}     & \multicolumn{1}{l}{0.49}     & \multicolumn{1}{l}{0.17}     & \multicolumn{1}{l|}{0.49}      \\ 
		\hline
		ResNet50~\cite{he2016deep} 
		& 2048           & 10.6         & \multicolumn{1}{l}{0.29}     & \multicolumn{1}{l}{0.36}     & \multicolumn{1}{l}{0.32}     & \multicolumn{1}{l}{0.17}     & \multicolumn{1}{l|}{0.31}      \\
		ResNet50\_PCA\_128
		&128         & 12.5         & \multicolumn{1}{l}{0.32}     & \multicolumn{1}{l}{0.34}     & \multicolumn{1}{l}{0.33}     & \multicolumn{1}{l}{0.17}     & \multicolumn{1}{l|}{0.31}      \\
		ResNet50\_PCA\_512
		& 512        & 12.5         & \multicolumn{1}{l}{0.34}     & \multicolumn{1}{l}{0.35}     & \multicolumn{1}{l}{0.34}     & \multicolumn{1}{l}{0.20}     & \multicolumn{1}{l|}{0.33}      \\
		GSDP\_RNet\_128~(our)
		&128          & 9.6         & \multicolumn{1}{l}{\textbf{0.43}}     & \multicolumn{1}{l}{\textbf{0.69}}     & \multicolumn{1}{l}{\textbf{0.53}}     & \multicolumn{1}{l}{\textbf{0.26}}     & \multicolumn{1}{l|}{\textbf{0.52}}      \\
		GSDP\_RNet\_512~(our)
		& 512         & 9         & \multicolumn{1}{l}{0.34}     & \multicolumn{1}{l}{0.47}     & \multicolumn{1}{l}{0.40}     & \multicolumn{1}{l}{0.09}     & \multicolumn{1}{l|}{0.39}      \\
		
		\hline
	\end{tabular}
}
\end{table}

\begin{figure}[t!]
	\begin{center}
		\includegraphics[width=0.95\linewidth]{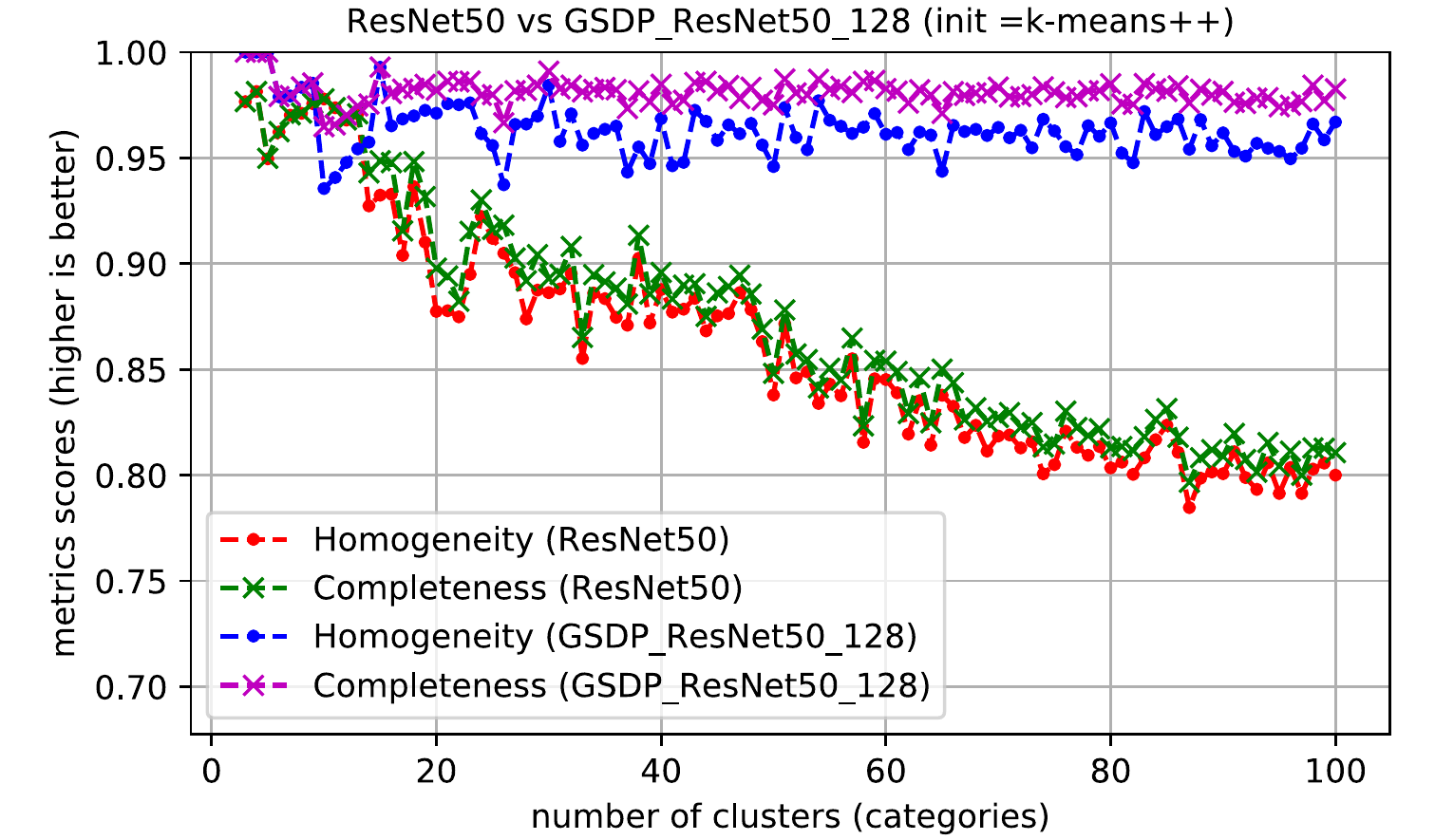}
		
	\end{center}
	 \vspace*{-\baselineskip}
	\caption{\textbf{K-Means metrics on ImageNet}. History of K-Means metrics reached by ResNet50 features\textit{ versus} our GSDP representation in the first $100$ categories of ImageNet dataset. 
	}
	\label{fig:Kmeans_metrics_Imagenet}
\end{figure}

Table~\ref{table:Kmeans_metrics} shows the results achieved by each global image descriptor on the $18$-th iteration of the experiments using K-Means clustering metrics: Homogeneity~(H), Completeness~(C), V-measure~(V), Adjusted Rand Index~(ARI), and Adjusted Mutual Information~(AMI). Figure~\ref{fig:Kmeans_metrics_Imagenet} depicts an example of K-Means metrics history achieved for ResNet50 features against our GSDP signatures in the first $100$ categories of the ImageNet dataset. 
The experiments show that as the data diversity of objects' images increases, our semantic GSDP encoding significantly outperforms another image global encoding in terms of cluster metrics in the ImageNet dataset. Furthermore, we conducted the same experiment in Coco (cross-dataset) to evaluate each image representation's performance and generalization ability on unseen data. 
Experiments showed that even when all image representations evaluated performed poorly in the Coco dataset, our GSDP representations performed best. The experiments show that the lowest dimensional GSDP representations (for each CNN model) were the ones that achieved the best size-performance trade-off.



		
	

%
\section{Conclusion}

In this paper, we introduced a Computational Prototype Model~(CPM) based on the foundations of Prototype Theory. Our approach provides another point of view for semantic representation of the internal structure of object categories. Our proposal retrieved some experimental psychology results to model some semantic properties of the object's image (\eg~typicality) that were still not analyzed by current prototype learning approaches.

We presented a straightforward Prototypical Similarity Layer~(PS-Layer), which uses the constraints of the CPM model to learn object categories, and it allowed the evaluation of the CPM model in classification tasks and transfer learning. The experiments carried out pre-computed semantic prototypes as prior knowledge, which did not update during the training process. Cross-dataset experiments showed that even under these unfavorable training conditions, the semantic information captured with the CPM model could be robust and achieve reasonable performance.

Furthermore, using the CPM model components (semantic prototype and semantic distance), we proposed a prototype-based description model~(GSDP) that introduces a new approach to the semantic description of objects' images. The GSDP descriptor built discriminatory signatures that semantically describe objects' images highlighting its most distinctive features within the category. Experiments in large image datasets showed that GSDP-descriptor is discriminative, small dimensioned, and can encode/preserves the semantic information of category members captured by the CPM model.

In summary, our experiments\footnote{All source code, prototypes datasets, GSDP tutorial and PS-Layer experiments examples will be  publicly available in the project page: https://www.verlab.dcc.ufmg.br/global-semantic-description/.} showed that the CPM model could encapsulate prominent semantic features of the object category in our semantic prototype representation, features that allow simulating the central and peripheral meaning of the category. Moreover, we showed that the semantic distance metric proposed in this article could simulate semantic relationships in terms of visual typicality between category members. Our prototypical distance can be understood as an object's image typicality score in which our CPM model can capture the visual representativeness degree of the object. The experiments also showed that it is possible to build robust semantic entities using little data (only typical images).

\paragraph{Limitations and Future works}

The lack of data sets with annotations of the image's typicality prevented a more robust evaluation of our approach. This limitation also generated deep learning feature engineering (or post-processing) to evaluate our CPM-model since the end-to-end training wasn't feasible. Consequently, as future work, we intend to construct a new image dataset with typicality annotations according to the interpretation criteria of human beings. With this initial work, we intend to encourage the pattern recognition community to delve into how to capture the image's typicality, a semantic property known to influence the learning process but which, to date, is only a skill of human beings.

\paragraph{Acknowledgment}
This research was supported by funding from the Brazilian agencies CAPES, CNPq, and FAPEMIG.


\bibliography{reference}

\end{document}